\documentclass[11pt]{article}

\usepackage[table]{xcolor}
\usepackage{tablefootnote}
\usepackage{float}
\usepackage{latex/acl}
\usepackage{placeins}
\usepackage{times}
\usepackage{latexsym}
\usepackage{hyperref}

\usepackage[T1]{fontenc}

\usepackage[utf8]{inputenc}

\usepackage{microtype}

\usepackage{inconsolata}

\usepackage{graphicx}
\usepackage{subfigure}
\usepackage{listings}
\usepackage{makecell}

\usepackage{multirow}
\usepackage{diagbox} 
\usepackage[normalem]{ulem}

\usepackage{amsmath}
\usepackage{amsfonts}
\usepackage{amssymb}
\usepackage{mathtools}
\usepackage{booktabs}
\usepackage{xr}
\usepackage{hyperref}
\usepackage{graphicx}
\usepackage{subcaption}

\usepackage[normalem]{ulem}
\useunder{\uline}{\ul}{}
\usepackage[linesnumbered,ruled,vlined]{algorithm2e}

\usepackage[symbol]{footmisc}

\SetKwInOut{Input}{Input}
\SetKwInOut{Output}{Output}
\SetKwInOut{Parameter}{Parameter}

\title{VideoINSTA: Zero-shot Long Video Understanding via \\Informative Spatial-Temporal Reasoning with LLMs}

    \author{Ruotong Liao$^{1, 2,\footnotemark[1]\;}$, \; Max Erler$^{1, \footnotemark[1]\;}$, \; Huiyu Wang$^{3}$, \; Guangyao Zhai$^{2,3}$, \;
    \\ \textbf{Gengyuan Zhang}$^{1,2}$, \; \textbf{Yunpu Ma}$^{1,2,4, \footnotemark[2]\;}$, \; \textbf{Volker Tresp}$^{1,2}$ \\
$^{1}$LMU Munich $\;$ $\;$  $^{2}$Munich Center for Machine Learning (MCML) $\;$ \\
$^{3}$Technical University of Munich $\;$ $^{4}$Siemens AG \\ 
\texttt{ruotong.liao@outlook.com, \; cognitive.yunpu@gmail.com}\\
\texttt{volker.tresp@lmu.de}
}

\begin{document}
\maketitle

\begin{abstract}

In the video-language domain, recent works in leveraging zero-shot Large Language Model-based reasoning for video understanding have become competitive challengers to previous end-to-end models. However, long video understanding presents unique challenges due to the complexity of reasoning over extended timespans, even for zero-shot LLM-based approaches. The challenge of information redundancy in long videos prompts the question of what specific information is essential for large language models (LLMs) and how to leverage them for complex spatial-temporal reasoning in long-form video analysis. We propose a framework \textbf{VideoINSTA} , i.e.
\textbf{IN}formative \textbf{S}patial-\textbf{T}empor\textbf{A}l Reasoning
for zero-shot long-form video understanding.
\textbf{VideoINSTA} contributes 
(1) a zero-shot framework for long video understanding using LLMs; 
(2) an event-based temporal
reasoning and content-based spatial reasoning approach for LLMs to reason over spatial-temporal information in videos; 
(3) a self-reflective information reasoning scheme based on information sufficiency and prediction confidence while balancing temporal factors.
Our model significantly improves the state-of-the-art on three long video question-answering benchmarks: EgoSchema, NextQA, and IntentQA, and the open question answering dataset ActivityNetQA. Code is released \href{https://github.com/mayhugotong/VideoINSTA}{here}. 

\end{abstract}

\addtocounter{footnote}{1}
\footnotetext{\;Equal contribution.}

\addtocounter{footnote}{2}
\footnotetext{\;Corresponding author.}


\section{Introduction}
\label{sec: intro}

Large language models (LLMs) have demonstrated remarkable reasoning abilities, even in long-context situations~\cite{chen2024longlora, mao2023large, NEURIPS2022_8bb0d291}. These advancements have spurred interest in video reasoning. Previous works bridging video and text modalities depend on meticulously designed models suffering large-scale pretraining. This challenge is pronounced with videos, a data format characterized by a vast volume of information scaling with length. Consequently, these models exhibit limited generalizability across datasets and struggle to scale to long video within a single model~\cite{Sun_2019_ICCV, NEURIPS2022_00d1f03b}. 
More recent models have gradually integrated LLMs' reasoning abilities by introducing lightly tuned adaptation layers~\cite{NEURIPS2022_00d1f03b, zhang2023videollama, lin2023videollava}. However, they still struggle with the length of the videos. Recently, to avoid expensive training costs, early attempts have proposed a zero-shot solution by reasoning over semantic representations of video content using LLMs~\cite{zhang2023simple, wang2024videoagent, choudhury2023zero}. These approaches have become strong competitors to earlier end-to-end models. Nonetheless, long-form video understanding, which demands advanced reasoning over extended timespans, remains challenging even for LLM-based methods.

Even in light of these tryouts, many challenges remain unsolved: 
(1) \textit{Information Quality}. 
Videos contain vast information even with some redundancy due to minor visual changes. Identifying the most crucial piece of information and extracting it effectively is essential to enhance the quality of data within the context window manageable by LLMs. How can we achieve this extraction?
(2) \textit{Neglect of Spatial and Temporal Characteristics}. 
Videos inherently exhibit temporal and spatial characteristics. How can we effectively preserve and convey this spatial-temporal information to support LLM reasoning? Especially, how do LLMs process temporal dynamics in videos?
(3) \textit{Complexity of Reasoning with Unbalanced Information over Temporal Span}.
In long videos, the significance of information along the video temporal axis varies greatly. LLMs' implicit "intuition" to process all the information is insufficient. How do we develop an explicit reasoning algorithm for unbalanced information considering temporal factor?

To address these challenges, we propose a framework \textbf{VideoINSTA}, i.e. \textbf{IN}formative
\textbf{S}patial-\textbf{T}empor\textbf{A}l reasoning for zero-shot long-form video understanding, aiming to build a compound system extracting essential information from long-form videos -- 
leveraging spatial-temporal reasoning and temporal-aware self-reflective reasoning to handle complex information with LLMs.

\textbf{VideoINSTA} is a zero-shot framework for reasoning with LLMs, augmented with visual-language tools. 
First, this framework emphasizes \textit{event-based temporal reasoning} by proposing an automatic temporal segmentation method C-DPCKNN, which segments long videos into multiple events. 
Besides, it derives the global temporal information with the help of a unified temporal representation tool UniVTG~\cite{qinghong2023univtg} and utilizes a temporal grounding scheme allowing the event to inherit the local temporal information.
Second, this framework emphasizes \textit{content-based spatial reasoning} by improving video captions with various visual-language captioning tools to extract richer spatial information. 
Specifically, event captioning is compensated by object detection and action caption as spatial information. A follow-up summarization serves as implicit spatial reasoning in a chain-of-thought manner.
Third, this framework proposes \textit{Iterative Information Reasoning} with LLMs, 
which iteratively merges the temporal and spatial information derived in the previous stages based on the self-evaluation of LLMs on the information sufficiency and prediction confidence.
    
Experiments have showcased remarkable improvements in existing long-form video question-answering tasks compared to end-to-end video-language models as well as other zero-shot LLM-based video understanding compound systems. Besides, VideoINSTA handles long videos with an average length of 3 minutes and is easily extensible for longer videos in a zero-shot manner.
This framework also shows excellent results both on multi-choice and open-question answering tasks.
The main contributions are summarized as follows:
\begin{itemize}
    \item \textbf{VideoINSTA: A zero-shot framework for long-form video understanding with state-of-the-art performance.} We propose a new zero-shot and extensible framework based on LLMs augmented with visual-language tools.
    \item \textbf{Spatial-temporal reasoning on videos with LLMs.} We propose event-based temporal reasoning and content-based spatial reasoning with LLMs utilizing extracted spatial-temporal information for understanding long-form videos.
    \item \textbf{Self-reflective information reasoning with LLMs considering temporal factors.} Our framework contributes to an iterative reasoning scheme for LLMs to merge and reason on the spatial-temporal information in a self-reflective manner while considering the temporal factors.
\end{itemize}


\section{Related Works}
\label{sec: rel_works}

\paragraph{Video Question Answering with LLMs}

Long video question answering involves predicting the correct answer given videos and queries, and optional multi-choice options. With advancements in LLMs and their long-context reasoning abilities, video understanding using LLMs has been explored in various works~\cite{xu2023retrieval, maaz2023video, jin2024chat, yu2024self, lin2023video, zhang2023video, huang2024vtimellm, wang2023vamos}. However, even with lightly tuned adaptation layers, scaling training costs increase significantly with video length. Recently, zero-shot methods like \cite{wang2022language} use image descriptors for video understanding tasks. Besides, LLoVi~\cite{zhang2023simple} and VideoAgent~\cite{wang2024videoagent}, which use extensive captioning and iterative keyframe selection respectively, have aimed to achieve training-free video understanding. Additionally, works such as ProViQ~\cite{choudhury2023zero} and MoReVQA~\cite{min2024morevqa} investigate zero-shot understanding using neuro-symbolic programming. 
LangRepo~\cite{kahatapitiya2024language} has a structured language repository to maintain textual video representations. TraverLER~\cite{shang2024travelermultilmmagentframework} iteratively gathers relevant information from keyframes with multiple LLMs and VideoTree~\cite{wang2024videotreeadaptivetreebasedvideo} is an extension of LLoVi with tree-based information searching scheme.
Unlike these approaches, we allows LLMs to directly reason on extracted spatial-temporal information without neuro-symbolic programming.

\paragraph{Spatial-Temporal Reasoning on Video}

Spatial-temporal reasoning in video has been a topic of continuous discussion~\cite{hussein2019videograph, wang2021supervoxel, xiao2023contrastive, wu2021spatial, zhu2022relational, jin2024chat, li2022invariant, xiao2022video, xiao2024can,zhai2020poseconvgru} due to the dual characteristics of video data. Most previous approaches compress information and perform reasoning within the embedding space. Additionally, recent works have highlighted LLMs' capabilities in temporal~\cite{tan2023towards,han-etal-2023-ecola, yuan2024back, liao-etal-2024-gentkg, ding-etal-2024-zrllm, xiong2024large} and spatial reasoning~\cite{ranasinghe2024learning, wu2024symbol, ko2023large, sharma2024vision, yamada2023evaluating, wu2024mind}. However, applying LLMs' spatial-temporal reasoning abilities to video remains underexplored. Our work innovatively harnesses these abilities, augmenting them with spatial-temporal reasoning methods as tools, to effectively analyze long-form videos both spatially and temporally.


\section{VideoINSTA: Informative Spatial-Temporal Reasoning with Large Language Models}
\label{sec: videoinsta}

\externaldocument{5_exp}

\begin{figure*}[ht]
    \centering 
    \includegraphics[width=0.95\linewidth]{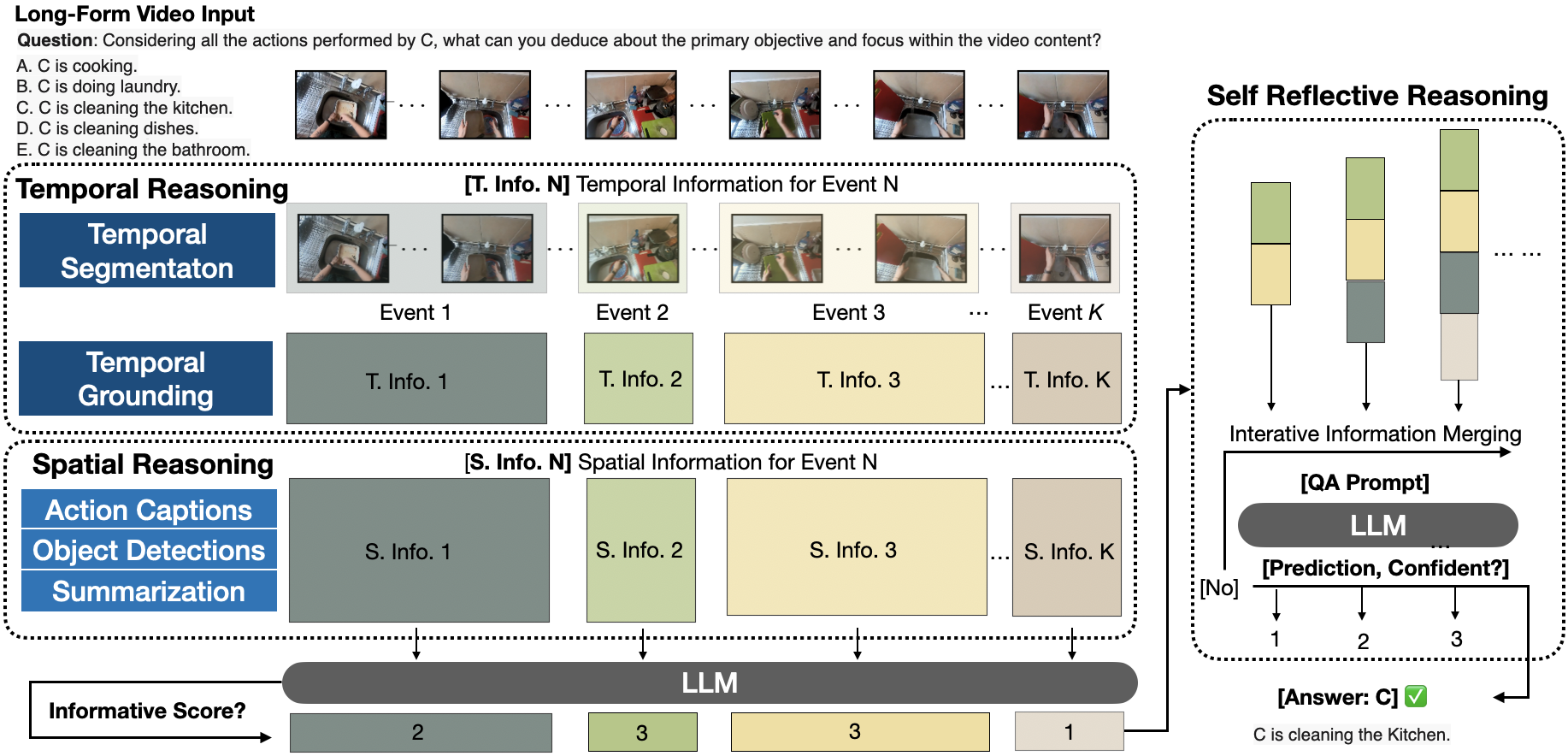}
    \caption{Framework of \textbf{VideoINSTA}. 
    VideoINSTA consists of three phases. (1) Event-based Temporal Reasoning. Temporal Segmentation parses the video into events via proposed C-DPCKNN clustering,
    and Temporal Grounding derives semantic temporal information inherited from the global relevance of each event. 
    (2) Content-based Spatial Reasoning. Action Captions are derived for each clip by video captioners as basic spatial information. 
    Compensated with Object Detections, the spatial information is summarized to derive query-focused spatial information. 
    (3) Self-reflective Information Reasoning. The previously derived spatial-temporal information is merged according to their
    information sufficiency in descending order and the LLM performs multi-round predictions after information merging 
    until it comes to a confident self-evaluation.}
    \label{fig: main}
\end{figure*}

In this section, we explain our \textbf{VideoINSTA} framework shown in Figure \ref{fig: main} following its three-phase methodology: event-based temporal reasoning, content-based spatial reasoning, and self-reflective information reasoning with LLMs.

\subsection{Event-based Temporal Reasoning}
\label{subsec: event_based}

\begin{figure}[h]
    \centering
    \includegraphics[width=1.0\linewidth]{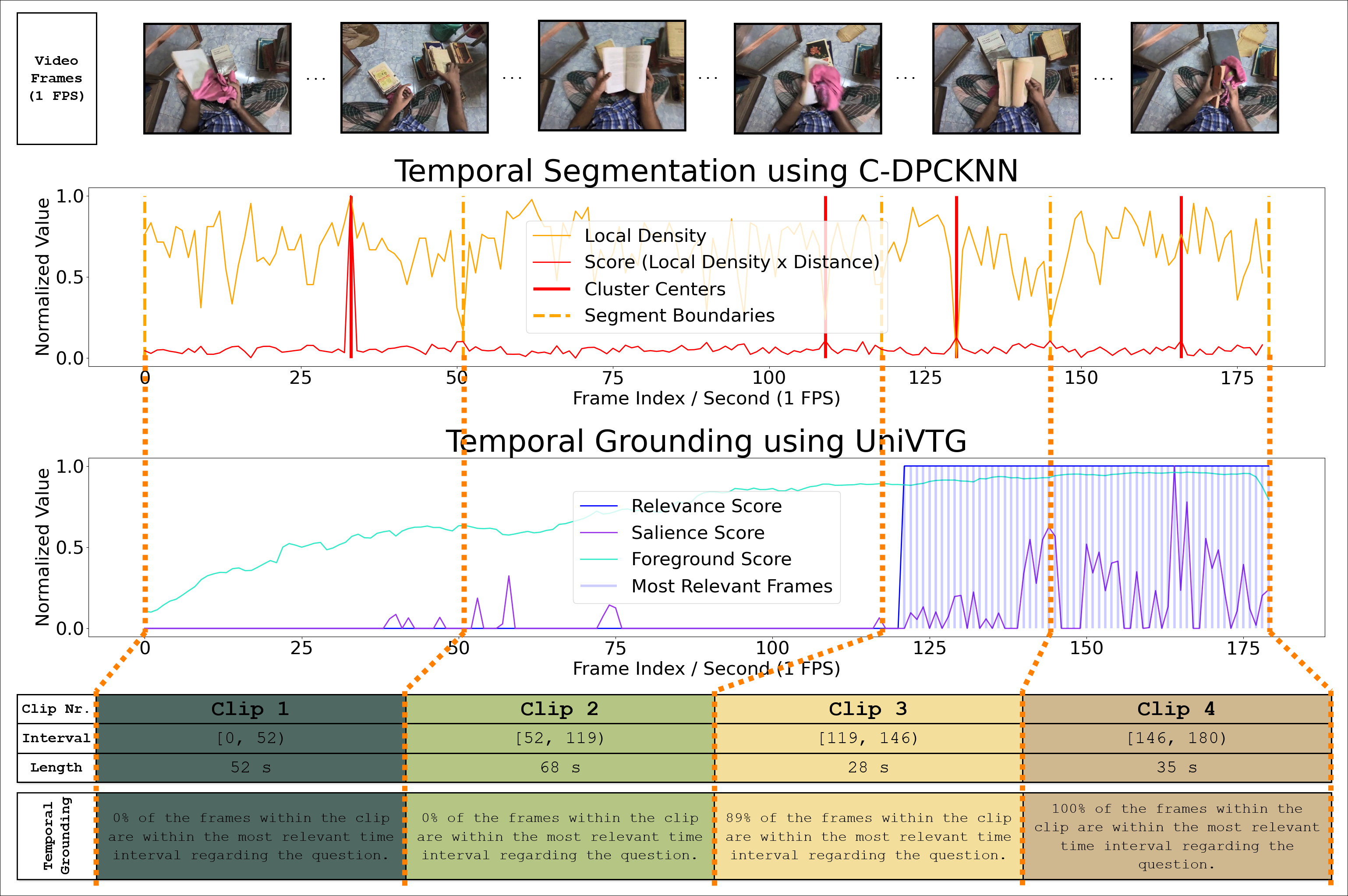}
    \caption{Illustration of Temporal Reasoning in \textbf{VideoINSTA}. 
    In Temporal Segmentation, the proposed C-DPCKNN sets clear borders with minimum density peaks. 
    In Temporal Grounding, each event inherits the global relevance information derived from UniVTG according to these borders. 
    The inherited local temporal information is transformed into semantic prompts, empowering temporal reasoning in \textbf{VideoINSTA}. }
    \label{fig: tem}
\end{figure}

The event-based temporal reasoning, as shown in Figure \ref{fig: tem}, consists of two sequential sub-steps differentiated by whether the query $Q$ is a known, specifically, query-agnostic temporal segmentation and query-aware temporal grounding. 

\subsubsection{Query-agnostic Temporal Segmentation}
\label{subsubsec: query_unaware}

KNN~\cite{guo2003knn} Clustering has been a widely used algorithm for temporal segmentation for separating event clips in video. 
For example, \cite{zhou2024streaming} utilizes KNN and ChatUniVi~\cite{jin2024chat} utilizes DPCKNN~\cite{du2016study},
a density-based clustering algorithm to merge frames belonging to the same events. 
However, these methods are designed specifically for embedding-based reasoning. 
They share a common fallback that frames or even tokens belonging to the same cluster scatter across the video span, causing blended boundaries between events, and frames from different events are interleaved thus not fulfilling the temporal order.
Therefore, we propose a \textit{consecutive} clustering algorithm \textbf{C-DPCKNN} for automatic event parsing on videos with clear boundaries. 

\paragraph{Event Center} Given a $i^{th}$ frame in a video, we first use the vision encoder of CLIP~\cite{radford2021learning} 
to provide its visual tokens $\boldsymbol{Z}=\left\{z_i\right\}_{i=1}^L$, 
where $L$ is the number of visual tokens within each frame.
Then we apply mean-pooling over all tokens to obtain the frame-level representation $f_i$. 
Specifically, we first compute the local density $\rho_mi$ as Eq. \ref{eq: rho}. 
Then we compute the distance index $\delta_i$ as Eq. \ref{eq: delta} of each frame $f_i$. 
We set frames with the highest $\rho_i \times \delta_i, i \in \left[1,2, \dots, M\right]$ as cluster centers, where $M$ is the total sampled frames in a video.
\begin{equation} 
\begin{gathered} \label{eq: rho}
\rho_i=\exp \left(-\frac{1}{K} \sum_{z_k \in \operatorname{KNN}\left(z_i, \boldsymbol{Z}\right)}\left\|z_k-z_i\right\|^2\right)\end{gathered}
\end{equation}

\begin{equation}\begin{gathered} \label{eq: delta}
\delta_i= \begin{cases}\min _{j: \rho_j>\rho_i}\left\|z_j-z_i\right\|^2, & \text { if } \exists j \text { s.t. } \rho_j>\rho_i \\ \max _j\left\|z_j-z_i\right\|^2, & \text { otherwise. }\end{cases}
\end{gathered}\end{equation}

\paragraph{Event Clustering} Given $K$ cluster centers, we cluster consecutive frames in both, forward and backward directions. We deprecate setting other frames directly to their nearest cluster center based on Euclidean distances of the embeddings which causes interleaved event frames and blurred boundaries that are counterintuitive to how events are separated and sequenced in an untrimmed video. Instead, we set the event boundary according to the critical points with the $K-1$ minimum density values, i.e. minimum density peaks $\boldsymbol{\Delta}=\left\{\delta_i\right\}_{i=1}^{K-1}$, indicating drastic changes in the frame content and denote the set of indexes of the frames in the cluster as $E$. We treat each cluster as a critical event and parse the events consistent with the frame order.

\paragraph{Event Segmentation} To set clear boundaries for each event, we store the indexes of boundary frames with $K-1$ minimum density peaks as $\boldsymbol{\mathcal{I}}=\left\{I_i\right\}_{i=1}^{K-1}$ to set the event set $\boldsymbol{\mathcal{E}}=\left\{E_i\right\}_{i=1}^K$ with respective starting and ending boundaries $\left\{(0,I_1),\dots,(I_{K-1},I_{EOV})\right\}_{i=1}^{K-1}$, $I_{EOV}$ denotes the ending index of video. The video is then parsed into respective event clips.

\subsubsection{Query-agnostic Temporal Grounding}
\label{subsubsec: query_aware}

Aside from automatic query-agnostic temporal segmentation, we introduce query-aware temporal grounding -- providing semantic temporal representations to support richer informative reasoning.

\paragraph{Global Temporal Relevance Derivation} We first derive the initial global temporal information, specifically, the relevance of the whole video given the query, with the help of the zero-shot unified video-language temporal grounding model UniVTG~\citep{qinghong2023univtg}. Given a video $V$ and a question query $Q$, UniVTG divides the original $V$ into fine-granular clips $V=\left\{v_i\right\}_{i=1}^{L_v}$ and evaluates each $v_i$ with triple evaluators $(f_i, b_i, s_i)_{i=1}^{L_v}$, where $L_v$ is the number of fine-grained clips. $s_i \in \left[0,1\right]$ are continuous salience scores determining the relevance between the visual content of the video and the query $Q$ spanning from totally irrelevant to highly correlated; $f_i$ are the foreground indicators for query-based moment retrieval, and $b_i$ are the boundary intervals for moment localization. 

\paragraph{Local Temporal Relevance Inheritance} 
As UniVTG derives global temporal relevance information for the whole video, we propose \textit{Local Inheritance} which assigns query-aware global temporal relevance information to the automatically and query-agnostic parsed event clips $\boldsymbol{\mathcal{E}}=\left\{E_i\right\}_{i=1}^K$ as local temporal relevance information. Specifically, a boundary-based inheritance scheme is performed. We rank fine-grained clips $\left\{v_i\right\}_{i=1}^{L_v}$ with predicted boundaries $\left\{b_i\right\}_{i=1}^{L_v}$ based on their $\left\{f_i\right\}_{i=1}^{L_v}$ probabilities and returns the Top-$k$ clips as query-aware moment retrieval predictions and return their boundaries $\left\{b_i\right\}_{i=1}^k$ given a question $V$ and a query $Q$. Then, we take boundary intersections between $\boldsymbol{\mathcal{I}}$ and $\left\{b_i\right\}_{i=1}^k$ and calculated the percentage of $\left\{b_i\right\}_{i=1}^k$ allocated in each event $E_i$. The relevance percentage is translated into semantic representations for LLMs to reason. Hence, the temporal information is transformed as prompt $\boldsymbol{\mathcal{P}}^t$.

\subsection{Content-based Spatial Reasoning}
\label{subsec:content_based}

\begin{figure}
    \centering
    \includegraphics[width=0.95\linewidth]{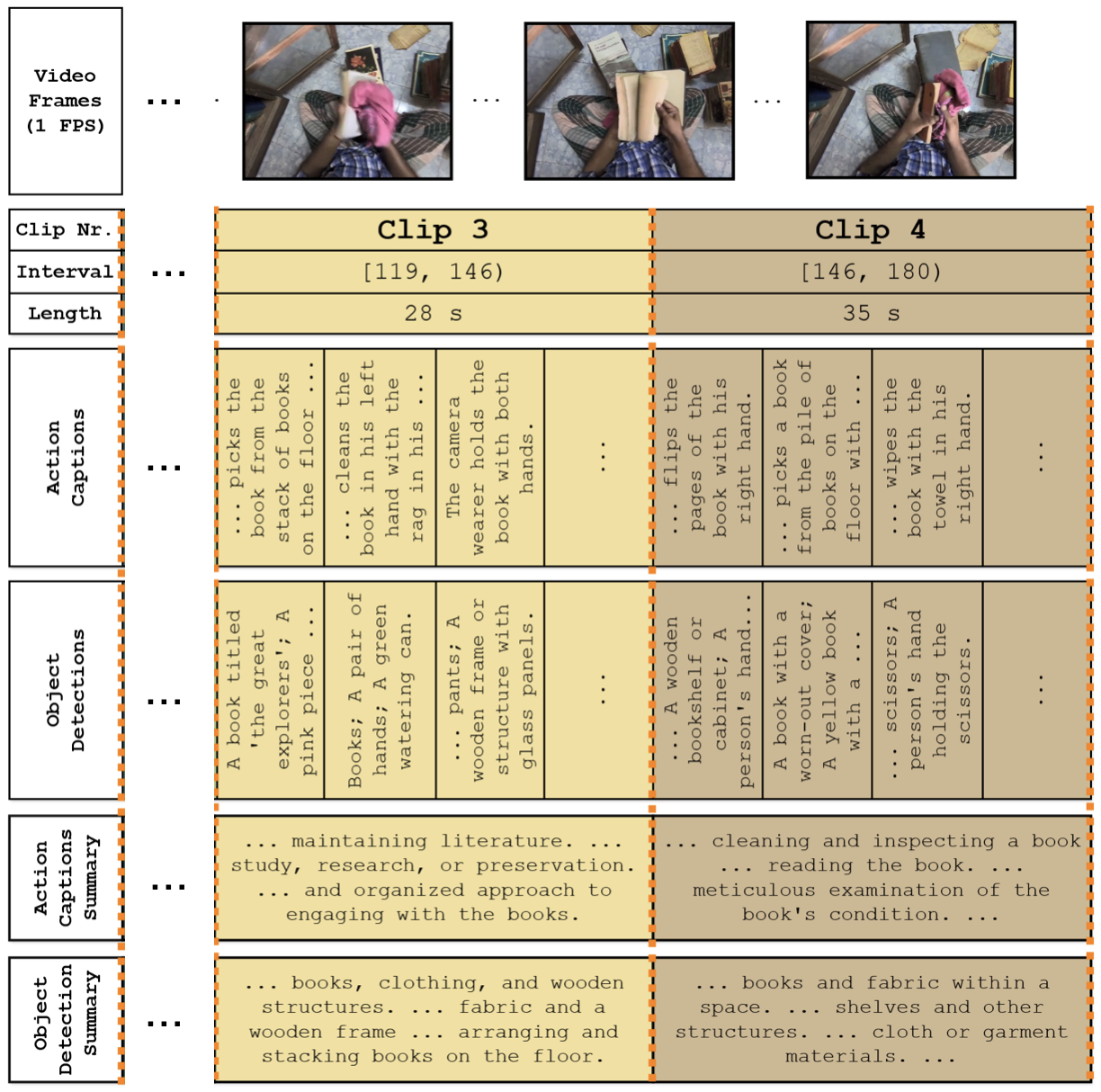}
    \caption{Spatial Reasoning in \textbf{VideoINSTA}. }
    \label{fig: spatial}
\end{figure}

The second phase of VideoINSTA contributes spatial reasoning with spatial information extraction. A common bottleneck from previous works on LLM-based video understanding is the redundant and inaccurate information in describing videos, especially overloading the LLMs' context window when processing long videos. It is necessary to address the importance of information density of the spatial information for LLMs to reason, especially for long-form videos. VideoINSTA shows that actions and objects occurring in the videos are the most crucial components. For each event clip in $\boldsymbol{\mathcal{E}}=\left\{E_i\right\}_{i=1}^K$, we derive informative prompts with action captions $\boldsymbol{\mathcal{P}}^a=\left\{P^a_i\right\}_{i=1}^K$ and object captions $\boldsymbol{\mathcal{P}}^o=\left\{P^o_i\right\}_{i=1}^K$, detailed as follows.

\subsubsection{Action Captioning}
\label{subsubsec: ac_cap}

We leverage generative visual-language models (VLMs) to convert the video context to language descriptions. To ensure zero-shot quality of the extracted spatial information and as a fair comparison to other approaches, we utilize LaViLa~\cite{zhao2023learning} -- pre-trained on Ego4D dataset~\cite{grauman2022ego4d}, following ~\cite{zhang2023simple} -- on ego-centric videos, to create automatic video narrations. The auto-generated narrations densely cover long videos while reserving temporal synchronization of the visual information and descriptions of the video actions within the event clip. For exo-centric videos, we follow ~\cite{wang2024videoagent} utilizing CogAgent ~\cite{hong2024cogagent} to provide descriptions of the sequential video frames with a special focus on events and actions, denoted as $\boldsymbol{\mathcal{P}}^a = \left\{P^a_i\right\}_{i=1}^K$, as in Appendix \ref{app: prompt}.

\subsubsection{Object Detections}
\label{subsubsec: ob_det}

Spatial awareness enhances reasoning by incorporating structural and contextual object descriptions of an image~\cite{chen2023shikra, ranasinghe2024learning}. We leverage the high-fidelity VLM CogAgent~\cite{hong2024cogagent} to extract objects from video frames as interactive subjects, aiding LLMs' spatial understanding. The VLM identifies a fixed number of prominent objects per frame. To maintain temporal consistency within an event clip, objects are sequentially stored as semantic representations (Fig.~\ref{fig: spatial}) for LLM reasoning, denoted as $\boldsymbol{\mathcal{P}}^o = \left\{P^o_i\right\}_{i=1}^K$, as in Appendix \ref{app: prompt}.

\subsubsection{Query-dependent Summarization}
\label{sss: FST}

Given a query, we prompt the LLMs to get a query-based summarization of the spatial information. The query-based summarization serves as an implicit Chain-of-Thought~\citep{wei2023chainofthought} for LLMs to reason over the spatial information, focusing on the query about long clips. The summarization step $\boldsymbol{\mathcal{P}}_s=\left\{P^s_i\right\}_{i=1}^K = \left\{(sum_{\mathcal{LLM}}(P^a_i,Q),sum_\mathcal{LLM}( P^o_i, Q))\right\}_{i=1}^K $ contains action summarizations focusing on event information and object summarizations focusing on environment information, as in Appendix \ref{app: prompt}.

\subsection{Informative Reasoning with Self-Reflection}
\label{subsubsec: self_refl}

\begin{algorithm}[t]
\SetAlgoLined
\SetArgSty{textnormal}
\tiny 
\caption{VideoINSTA} \label{alg: self-ref}
\Input{Video $V$, Question $Q$, Options $\{ o_0, o_1, o_2, o_3, o_4 \}$}
\Parameter{Number of segments $K \in \mathbb{N}^+$}
\Output{Final Prediction $answer \in \{ o_0, o_1, o_2, o_3, o_4 \}$}
  $V' \leftarrow \emptyset$\tcp*{for clip descriptions and informative scores}
  $\boldsymbol{\mathcal{E}} \leftarrow \mathtt{temporal\_segmentation(}V, K\mathtt{)}$\;
  $T \leftarrow \mathtt{temporal\_grounding(}V, Q\mathtt{)}$\;
  $A \leftarrow \mathtt{action\_captions(}V\mathtt{)}$\;
  $O \leftarrow \mathtt{object\_detections(}V\mathtt{)}$\;

  \For{$E_i \in \boldsymbol{\mathcal{E}}$}{
    $P^a_{i} \leftarrow \mathtt{inherit(}A, E_i\mathtt{)}$\;
    $P^o_{i} \leftarrow \mathtt{inherit(}O, E_i\mathtt{)}$\;
    $P^t_{i} \leftarrow \mathtt{inherit(}T, E_i\mathtt{)}$\;
    
    $P^{sa}_i \leftarrow \mathtt{summarize(}P^a_{i}, Q\mathtt{)}$\;
    $P^{so}_i \leftarrow \mathtt{summarize(}P^o_{i}, Q\mathtt{)}$\;
  
    $P_i \leftarrow (P^a_{i}, P^o_{i}, P^t_{i}, P^{sa}_i, P^{so}_i)$\;


    $S^I_i\leftarrow \mathtt{informative\_eval(}P_i, Q, (o_0, o_1, o_2, o_3, o_4)\mathtt{)}$\;
    $V'\mathtt{.insert(}(P_i, S^I_i)\mathtt{)}$\tcp*{$i$-th clip description and info score}
  }
  $V'' \leftarrow \mathtt{sort\_descending(}V', \mathtt{key}=V'\mathtt{.}S_I\mathtt{)}$\tcp*{by info scores}
  $L \leftarrow \emptyset$\tcp*{for merged clip descriptions without info scores}
 
  \For{$E_i \in V''$}{
    $P_i, S^I_i \leftarrow E_i$\;
    $L\mathtt{.insert(}P_i\mathtt{)}$\;
    \If{$i \neq |V''| - 1$ $\mathtt{and}$ $S^I_{(i+1)} = 3$}{
      $\mathtt{continue}$\;
    }
    \Else{
      $L' \leftarrow \mathtt{sort\_temporally(}L\mathtt{)}$\;
      $P_{L'} \leftarrow \mathtt{concatenate(}L'\mathtt{)}$\;
      $answer, prompt, completion \leftarrow \mathtt{QA(}P_{L'}, Q, (o_0, o_1, o_2, o_3, o_4)\mathtt{)}$\;
      $S^C_i \leftarrow \mathtt{self\_reflect(}prompt, completion\mathtt{)}$\;
      \If{$S^C_i = 3$}{$\mathtt{break}$\;}
    }
  }

\Return{$answer$\;}

\end{algorithm}

Inspired by Reflexion~\cite{NEURIPS2023_1b44b878}, the third phase of VidoeINSTA proposes a self-reflective information reasoning scheme -- with LLMs to reason on spatial-temporal information collected in the previous stages.  
Particularly, we balance between information sufficiency and the temporal order. 
Two evaluation scores are defined as intermediate metrics in our algorithm. 

\textbf{Informative Score.}
The LLM is required to generate an Informative Score $S_I=\left\{S^I_i\right\}_{i=1}^K \in \left[1,2,3\right]$ for each clip indicating $\left[\text{not sufficient},\text{marginal sufficient},\text{sufficient}\right]$, which is an initial evaluation of the information sufficiency of the prompts derived in previous stages. 

\textbf{Confidence Score.} The LLM is required to generate a Confidence Score $S_C=\left\{S^C_i\right\}_{i=1}^K \in \left[1,2,3\right]$ for each question-answering round indicating $\left[\text{not confident},\text{marginal confident},\text{very confident}\right]$, which is a self-evaluation of the answer prediction.

\textbf{Self-reflective reasoning.} The algorithm shown in Alg. \ref{alg: self-ref} starts with an initial evaluation step for the LLM to derive an informative score for each clip. Then, the informative states are sorted in descending order according to their informative scores and maintained in a list. Within the same informative level, the prompts are ordered temporally. Then, the algorithm performs a multi-round self-reflective scheme, specifically merging informative clips and evaluating the question-answering confidence. In the first round, sufficient informative states are merged and prompted to the LLM for question-answering. Then, the LLM is required to derive a confidence score. If the LLM is not confident enough about its prediction, a further clip with a lower informative score is merged into the state which gets temporally re-ordered. The alternating merge-and-evaluate scheme ends until all clips are merged or the prediction confidence reaches the top value. The VideoINSTA is detailed in Alg. \ref{alg: self-ref} and on the right of Figure. \ref{fig: main}.


\section{Extensibility of the Framework}

\paragraph{Extensible API tools}
VideoINSTA is a general framework for informative spatial-temporal
reasoning on videos and maintains the extensibility to improve both, the temporal reasoning and spatial
reasoning phases by acquiring informative prompts from different expert tools through APIs. 
For example, expert temporal segmentation models can be utilized for better
event parsing in the temporal reasoning phase in VideoINSTA. Expert spatial models
like high-fidelity captioning models and object 
detectors can provide more 
accurate informative prompts for the spatial reasoning phase.
\paragraph{Open Question Answering}
Apart from single-choice question answering, VideoINSTA can also be easily 
adapted to open question answering. We tested VideoINSTA on AcitivityNet-QA
~\cite{yu2019activitynetqa}, which is a dataset for open-ended question answering over complex web videos. 
Following~\cite{Maaz2023VideoChatGPT}, we also conduct evaluation in a zero-shot 
manner, employing LLM-assisted evaluation to assess the predictions' accuracy
 of VideoINSTA.


\section{Experimental Setup}
In this section, we describe the experimental setup of the VideoINSTA framework. We present quantitative results and a qualitative analysis on the EgoSchema~\cite{mangalam2024egoschema},
Next-QA~\cite{xiao2021next}, and Intent-QA~\cite{li2023intentqa} benchmarks.

\textbf{EgoSchema} EgoSchema is a benchmark for long-form video understanding, featuring 5,000 single-choice questions derived from egocentric videos. A distinctive feature of this dataset is the length of its videos, each
lasting 180 seconds. EgoSchema comprises only a test set, with a subset of 500 questions having available labels. 

\textbf{NextQA} The NExT-QA dataset includes 5,440 natural videos that feature object interactions in daily life, accompanied by 48,000 single-choice questions. The average length of the video is 44 seconds. In line with standard practices, our zero-shot evaluation is focused on the validation set.

\textbf{IntentQA} IntentQA focuses on intent reasoning. It contains 4,303 videos and 16K single-choice question-answer pairs focused on reasoning about people’s intent in the video. The videos are more than 44 seconds in average length. We perform a zero-shot evaluation on the test set. 

\paragraph{Evaluation Metrics}
Since each dataset features single-choice questions and VideoINSTA generates option predictions directly, we utilized accuracy as the evaluation metric.

\paragraph{Baselines}
The baselines include recent representative LLM-based
zero-shot video understanding methods -- 
including LLoVi, VideoAgent, ProViQ and MoReVQA -- and 
other baselines include supervised end-to-end models, see Table \ref{tab: main_results}.

\paragraph{Experiment Design}
To comprehensively analyze VideoINSTA, there are two research questions. \textbf{RQ1}: How is the performance of the proposed VideoINSTA framework compared to the existing end-to-end models and LLM-based compound systems? \textbf{RQ2}: How do the components of the VideoINSTA affect its effectiveness? 

\paragraph{Implementation Details}
Following LLoVi and VideoAgent, we utilize the LaViLa model re-trained on Ego4D, filtering out videos that overlap with EgoSchema to ensure zero-shot evaluation. 


\section{Experimental Results}
\externaldocument{10_appendix}

\definecolor{my_orange}{RGB}{252, 187, 164}
\definecolor{my_yellow}{RGB}{252, 252, 164}
\definecolor{gray_lvl_1}{RGB}{153, 153, 153}
\definecolor{gray_lvl_2}{RGB}{170, 170, 170}
\definecolor{gray_lvl_3}{RGB}{187, 187, 187}
\definecolor{gray_lvl_4}{RGB}{204, 204, 204}
\definecolor{gray_lvl_5}{RGB}{221, 221, 221}
\definecolor{gray_lvl_6}{RGB}{238, 238, 238}

\begin{table*}[h]
\SetArgSty{textnormal}
\small 
\begin{center}
        \begin{center}
            \resizebox{\textwidth}{!}{
            \begin{tabular}{ll|lll}
                \hline
                \multicolumn{2}{l|}{\cellcolor{gray_lvl_1} \diagbox{Method}{Dataset}}                                 & \multicolumn{1}{l|}{ \cellcolor{gray_lvl_3} EgoSchema}                 & \multicolumn{1}{l|}{ \cellcolor{gray_lvl_3} NExT-QA}             &  \cellcolor{gray_lvl_3} IntentQA \\ \hline
                \multicolumn{2}{l|}{\cellcolor{gray_lvl_3}Random Chance}                               & \multicolumn{1}{l|}{20.0}                    & \multicolumn{1}{l|}{20.0}              & 20.0   \\ \hline
                \multicolumn{2}{l|}{\cellcolor{gray_lvl_3}Supervised State-of-the-Art}                      & \multicolumn{3}{l}{}                                                                                 \\ \hline
                \multicolumn{2}{l|}{LongViViT ~\cite{papalampidi2023simple}}                                   & \multicolumn{1}{l|}{56.8}                      & \multicolumn{1}{l|}{-}                    & -      \\
                \multicolumn{2}{l|}{MC-ViT-L ~\cite{balažević2024memory}}                                    & \multicolumn{1}{l|}{62.6}   & \multicolumn{1}{l|}{65.0}                    & -      \\
            \hline
                \multicolumn{2}{l|}{\cellcolor{gray_lvl_3}Training-Free State-of-the-Art}                  & \multicolumn{3}{l}{}                                                                \\ \cline{1-2}
                \multicolumn{1}{l|}{\cellcolor{gray_lvl_5}LLM}                          & \cellcolor{gray_lvl_5}System       & \multicolumn{3}{l}{}                                                                                 \\ \hline
                \multicolumn{1}{l|}{PaLM-2~\citep{anil2023palm2technicalreport}}                       & MoReVQA ~\cite{ranasinghe2024understanding}     & \multicolumn{1}{l|}{51.7\footnotemark[2]} & \multicolumn{1}{l|}{69.2}              & -      \\ \hline
                \multicolumn{1}{l|}{FlanT5-3B~\citep{2020t5}}                       & \multicolumn{1}{|l|}{SeViLA ~\cite{yu2024self}}                                      & \multicolumn{1}{l|}{25.7}                      & \multicolumn{1}{l|}{63.6} & 60.9    \\ \hline
                \multicolumn{1}{l|}{\multirow{2}{*}{Mistral-7B~\citep{jiang2023mistral7b}}}   & LangRepo~\citep{kahatapitiya2024languagerepositorylongvideo}     & \multicolumn{1}{l|}{60.8}                      & \multicolumn{1}{l|}{54.6}                   & 53.8      \\
                \multicolumn{1}{l|}{} & MVU~\cite{ranasinghe2024understanding}     & \multicolumn{1}{l|}{60.3}                      & \multicolumn{1}{l|}{55.2}                   & -      \\ \hline
                \multicolumn{1}{l|}{Llama2-7B~\citep{touvron2023llama}}                    & LLoVi ~\cite{zhang2023simple}       & \multicolumn{1}{l|}{34.0}                    & \multicolumn{1}{l|}{-}                 & -      \\ \hline
                \multicolumn{1}{l|}{Llama2-13B~\citep{touvron2023llama}}                   & LLoVi ~\cite{zhang2023simple}       & \multicolumn{1}{l|}{40.4}                    & \multicolumn{1}{l|}{-}                 & -      \\ \hline
                \multicolumn{1}{l|}{\multirow{2}{*}{Llama2-70B~\citep{touvron2023llama}}}  & LLoVi ~\cite{zhang2023simple}       & \multicolumn{1}{l|}{50.6}                    & \multicolumn{1}{l|}{-}                 & -      \\
                \multicolumn{1}{l|}{}                             & VideoAgent ~\cite{wang2024videoagent}   & \multicolumn{1}{l|}{45.4}                    & \multicolumn{1}{l|}{-}                 & -      \\ \hline
                \multicolumn{1}{l|}{\multirow{1}{*}{GPT-3~\citep{brown2020languagemodelsfewshotlearners}}} & ViperGPT ~\cite{surís2023vipergpt}       & \multicolumn{1}{l|}{-}  & \multicolumn{1}{l|}{60.0}                 & -      \\ \hline
                \multicolumn{1}{l|}{\multirow{2}{*}{GPT-4V~\citep{openai2024chatgpt}}} & IG-VLM ~\cite{kim2024image}      & \multicolumn{1}{l|}{59.8}                      & \multicolumn{1}{l|}{68.6} & 64.2     \\
                \multicolumn{1}{l|}{}                             & GPT-4V ~\cite{balažević2024memory}   & \multicolumn{1}{l|}{63.5}   & \multicolumn{1}{l|}{-}                    & -   \\ \hline
                \multicolumn{1}{l|}{\multirow{2}{*}{Llama3-8B~\citep{dubey2024llama3herdmodels}}}   & LLoVi ~\cite{zhang2023simple} (ours) & \multicolumn{1}{l|}{\colorbox{my_yellow}{47.6}} & \multicolumn{1}{l|}{\colorbox{my_yellow}{46.6}}  & \colorbox{my_yellow}{48.9}   \\
                \multicolumn{1}{l|}{}                             & \textbf{VideoINSTA}   & \multicolumn{1}{l|}{\colorbox{my_orange}{52.6}}  & \multicolumn{1}{l|}{\colorbox{my_orange}{58.3}}  & \colorbox{my_orange}{53.0}   \\ \hline
                \multicolumn{1}{l|}{\multirow{7}{*}{ChatGPT-4~\citep{openai2024chatgpt}}}   & LLoVi ~\cite{zhang2023simple}       & \multicolumn{1}{l|}{61.2}                       & \multicolumn{1}{l|}{67.7}              & 64.0   \\
                \multicolumn{1}{l|}{}                             & AssistGPT ~\cite{gao2023assistgpt}  & \multicolumn{1}{l|}{-}                       & \multicolumn{1}{l|}{58.4}                   & -      \\
                \multicolumn{1}{l|}{}                             & VideoAgent ~\cite{wang2024videoagent}  & \multicolumn{1}{l|}{60.2}                    & \multicolumn{1}{l|}{ 71.3}              & -      \\
                \multicolumn{1}{l|}{}                             & VideoAgent ~\cite{fan2024videoagentmemoryaugmentedmultimodalagent}  & \multicolumn{1}{l|}{62.8}                    & \multicolumn{1}{l|}{ 70.8 }              & -      \\
                \multicolumn{1}{l|}{}                             & TraveLER ~\cite{shang2024travelermultilmmagentframework}   & \multicolumn{1}{l|}{ - }   & \multicolumn{1}{l|}{68.2}  &  -   \\
                \multicolumn{1}{l|}{}                             & VideoTree ~\cite{wang2024videotreeadaptivetreebasedvideo}   & \multicolumn{1}{l|}{\colorbox{my_orange}{\textbf{66.2}}}   & \multicolumn{1}{l|}{\colorbox{my_orange}{\textbf{73.5}}}  & \colorbox{my_yellow}{\underline{66.9}}  \\
                \multicolumn{1}{l|}{}                             & \textbf{VideoINSTA}   & \multicolumn{1}{l|}{\colorbox{my_yellow}{\underline{65.0}}}   & \multicolumn{1}{l|}{\colorbox{my_yellow}{\underline{72.3}}}  & \colorbox{my_orange}{\textbf{72.8}}  \\ \hline
                \multicolumn{1}{l|}{\multirow{5}{*}{ChatGPT-3.5~\citep{openai2024chatgpt}}} & LLoVi ~\cite{zhang2023simple}       & \multicolumn{1}{l|}{\colorbox{my_yellow}{58.8}}  & \multicolumn{1}{l|}{-}                 & -      \\
                \multicolumn{1}{l|}{} & ProViQ ~\cite{choudhury2023zero}       & \multicolumn{1}{l|}{57.1}                    & \multicolumn{1}{l|}{\colorbox{my_yellow}{63.8\footnotemark[3]}} & -      \\
                \multicolumn{1}{l|}{} & VideoAgent ~\cite{wang2024videoagent}       & \multicolumn{1}{l|}{ - }                    & \multicolumn{1}{l|}{48.8} & -      \\
                \multicolumn{1}{l|}{} & VideoTree ~\cite{wang2024videotreeadaptivetreebasedvideo}       & \multicolumn{1}{l|}{ 57.6 }                    & \multicolumn{1}{l|}{-} & -      \\
                \multicolumn{1}{l|}{}                             & \textbf{VideoINSTA}   & \multicolumn{1}{l|}{\colorbox{my_orange}{62.8}}  & \multicolumn{1}{l|}{\colorbox{my_orange}{67.9}} & \colorbox{my_orange}{64.4}   \\ \hline
            \end{tabular}
            }
        \end{center}
    \caption{Video Reasoning Results. The best accuracy (\%) is highlighted in \colorbox{my_orange}{orange} and the second best in \colorbox{my_yellow}{yellow} for each training-free (zero-shot or few-shot) method respectively. Note that we are strictly zero-shot without using in-context examples in our prompts. The best result among all methods is \textbf{bold} and the second best is \underline{underlined}. 
    } \label{tab: main_results}
\end{center}
\end{table*}

\subsection{Main Results}

\paragraph{Comparison with State-of-the-arts}

To answer the RQ1, our average results over multiple run from Table \ref{tab: main_results} achieve state-of-the-art performance, 
surpassing all types of existing end-to-end models, proprietary models, and zero-shot compound systems across three datasets.

Noticeably, \textbf{VideoINSTA with ChatGPT3.5 surpasses the other zero-shot LLM-based baselines LLoVi and VideoAgent 
with ChatGPT-4}. Our method demonstrates spatial-temporal informative reasoning to serve as the foundational framework 
for zero-shot video reasoning, opening a new state-of-the-art in the video question-answering domain. 

\paragraph{Open Question Answering}
We measure the accuracy by utilizing an LLM to evaluate
 the generated prediction by comparing it to the ground truth answer and assigning a 
 true or false value accordingly. Table \ref{tab: openqa} shows the results with Llama-3.
VideoINSTA achieves more than double the performance compared to the baseline LLoVi 
 with \textbf{151.3\%} relative improvement.


\begin{table}[!h]
    \centering
    \resizebox{0.48\textwidth}{!}{
    \begin{tabular}{c|c|c}
    \hline
         \cellcolor{gray_lvl_1} LLM & \cellcolor{gray_lvl_1} Model & \cellcolor{gray_lvl_1} Accuracy (\%)  \\ \hline
        \multirow{2}{*}{\parbox{3.5cm}{\centering Llama-3-8B-Instruct \\ \citep{llama3modelcard}}} & LLoVi & 14.75\\ \cline{2-3}
         & VideoINSTA & \textbf{37.06 (151.3\% $\uparrow$)}\\ \hline
    \end{tabular}}
    \caption{Accuracy performance of VideoINSTA on open question answering dataset ActivityNet-QA.}
 \label{tab: openqa}
\end{table}

\subsubsection{Ablation on Main Stage}
We undertake ablation studies on EgoSchema to evaluate the contribution of each phase in VideoINSTA
with three distinct variations: VideoINSTA w/o TA (without event-based temporal reasoning), VideoINSTA w/o S (without content-based spatial reasoning), and VideoINSTA w/o IN (without self-reflective information reasoning). We further investigate event-based temporal reasoning and the contribution of the query-unaware temporal segmentation (VideoINSTA w/o TA-Seg.) and the query-aware temporal inheritance (VideoINSTA w/o TA-Inhr.).
Figure \ref{fig: abl} concludes that all phases in the VideoINSTA framework
contribute to distinct performance improvements including the two sub-steps in the temporal reasoning.
The whole pipeline enables VideoINSTA to outperform existing methods. 

\subsection{Ablation on Temporal Reasoning}

\paragraph{Clustering in Temporal Segmentation} To evidently prove the effectiveness of our proposed C-DPCKNN, 
we conduct experiments on variants VideoINSTA w. TA-Seg. (Uniform), w. TA-Seg. (KNN), w. TA-Seg. (DPCKNN) and w. TA-Seg. (C-DPCKNN) on both EgoSchema and NExT-QA. 
The quantitative results of this comparison are illustrated in Figure \ref{fig: TA-segment}. The results validate that our proposed C-DPCKNN method for query-unaware temporal segmentation is superior to the other approaches. Additionally, the worse performance of Uniform, KNN, and DPCKNN highlights that improper segmentation can severely impact subsequent reasoning steps. We conclude that they have the same drawback of improper segmentation, further validating the effectiveness of C-DPCKNN.

 \begin{figure}[h]
     \centering
     \includegraphics[width=0.8\linewidth]{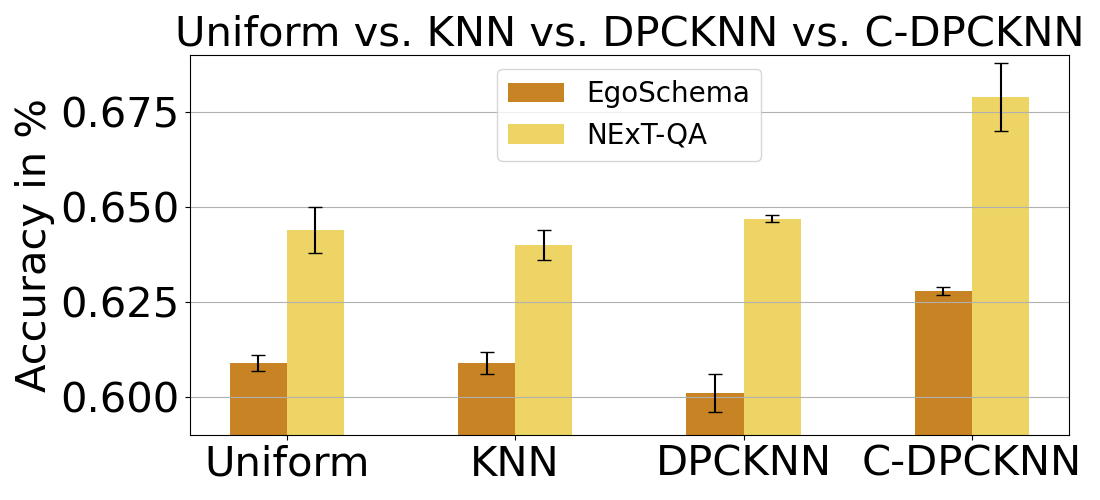}
     \caption{Ablation on different temporal segmentation of VideoINSTA methods.}
     \label{fig: TA-segment}
 \end{figure}

\begin{figure}[t]
    \centering
    \includegraphics[width=1.0\linewidth]{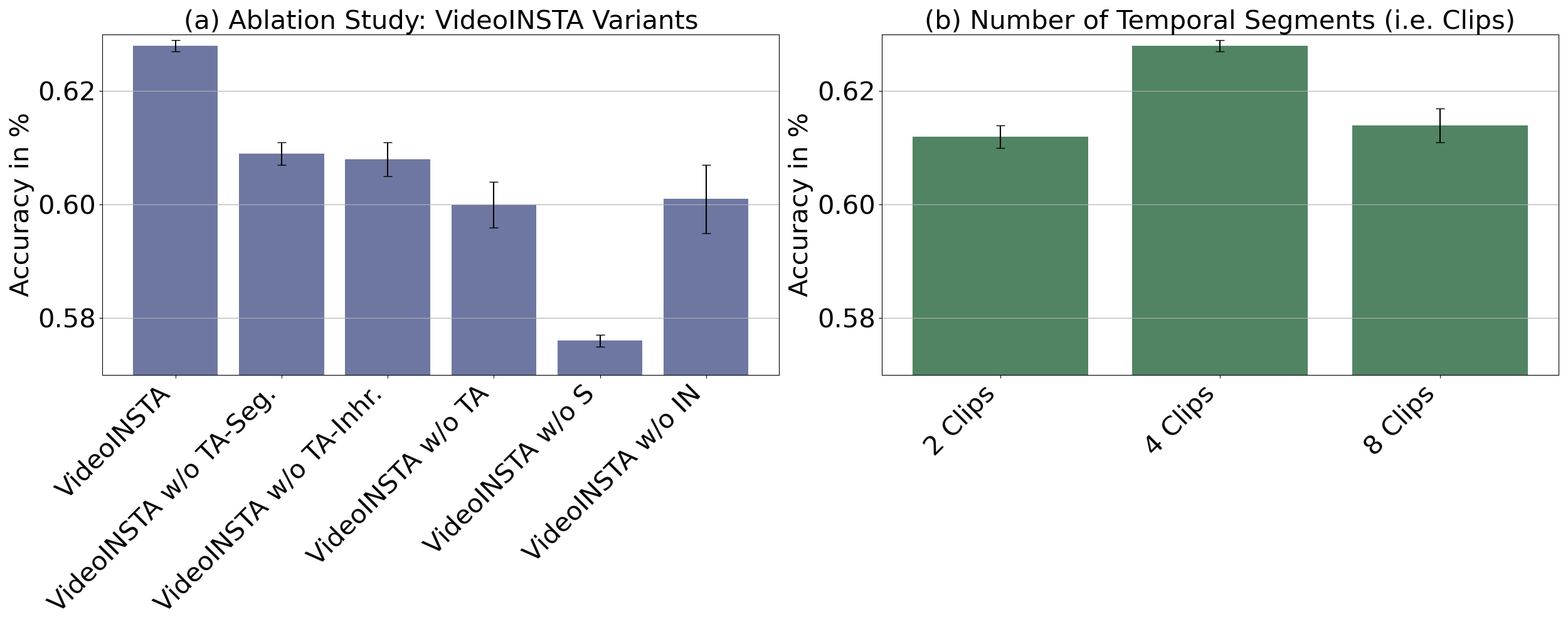}
    \caption{Ablation Studies on EgoSchema. 
    (a) All three phases contribute to \textbf{VideoINSTA}.
    (b) $K=4$ is the best empirical clustering number for EgoSchema.}
    \label{fig: abl}
\end{figure}

\paragraph{Number of Events in Temporal Segmentation}
To further explore the impact of C-DPCKNN in temporal segmentation within our temporal reasoning framework, we conducted a series of experiments on the EgoSchema dataset. We varied the number of event clips $K$ from the set $\{2, 4, 8\}$. For each configuration, we kept the implementation of other components in VideoINSTA consistent. Empirical results reveal an optimal critical value for the number of events $K$, as shown in Figure \ref{fig: abl}(b). EgoSchema videos are characterized by their uniform length of 3 minutes, with a high temporal certificate - a metric indicating the proportion of necessary informative segments to the total video duration. The empirical findings suggest that $K$ intuitively corresponds to the actual number of events observed in the videos. 

\subsection{Ablation on Spatial Stage}
\paragraph{Spatial Captioners}
We provide an ablation study over captioners comparing CogAgent vs. LLaVA-1.5~\cite{liu2023improvedllava} on NExT-QA, indicating that a better captioner leads to better information quality as CogAgent is a captioner with higher fidelity since it was especially designed for Graphical User Interface understanding
and navigation, which requires fine-granular perception. Therefore, CogAgent facilitates
better informativeness in tasks involving visual and
linguistics.

\begin{table}[!ht]
    \centering
    \resizebox{0.48\textwidth}{!}{
    \begin{tabular}{c|c|c}
    \hline
    \cellcolor{gray_lvl_1} LLM & \cellcolor{gray_lvl_1} Object Captioner & \cellcolor{gray_lvl_1} Accuracy \\ \hline
    \multirow{2}{*}{\parbox{3.5cm}{\centering ChatGPT-3.5 \\~\citep{openai2024chatgpt}}} & CogAgent & 0.679 \\ \cline{2-3}
                                 & LLaVA-1.5 & 0.628 \\ \hline
    \end{tabular}}
    \caption{Performance metrics for different captioners using ChatGPT3.5 on the NExT-QA dataset.}
    \label{tab:dataset_performance}
\end{table}
\subsection{Qualitative Analysis}

\paragraph{Event Segmentation with Clear Borders}

 \begin{figure}[t]
     \centering
     \includegraphics[width=1.0\linewidth]{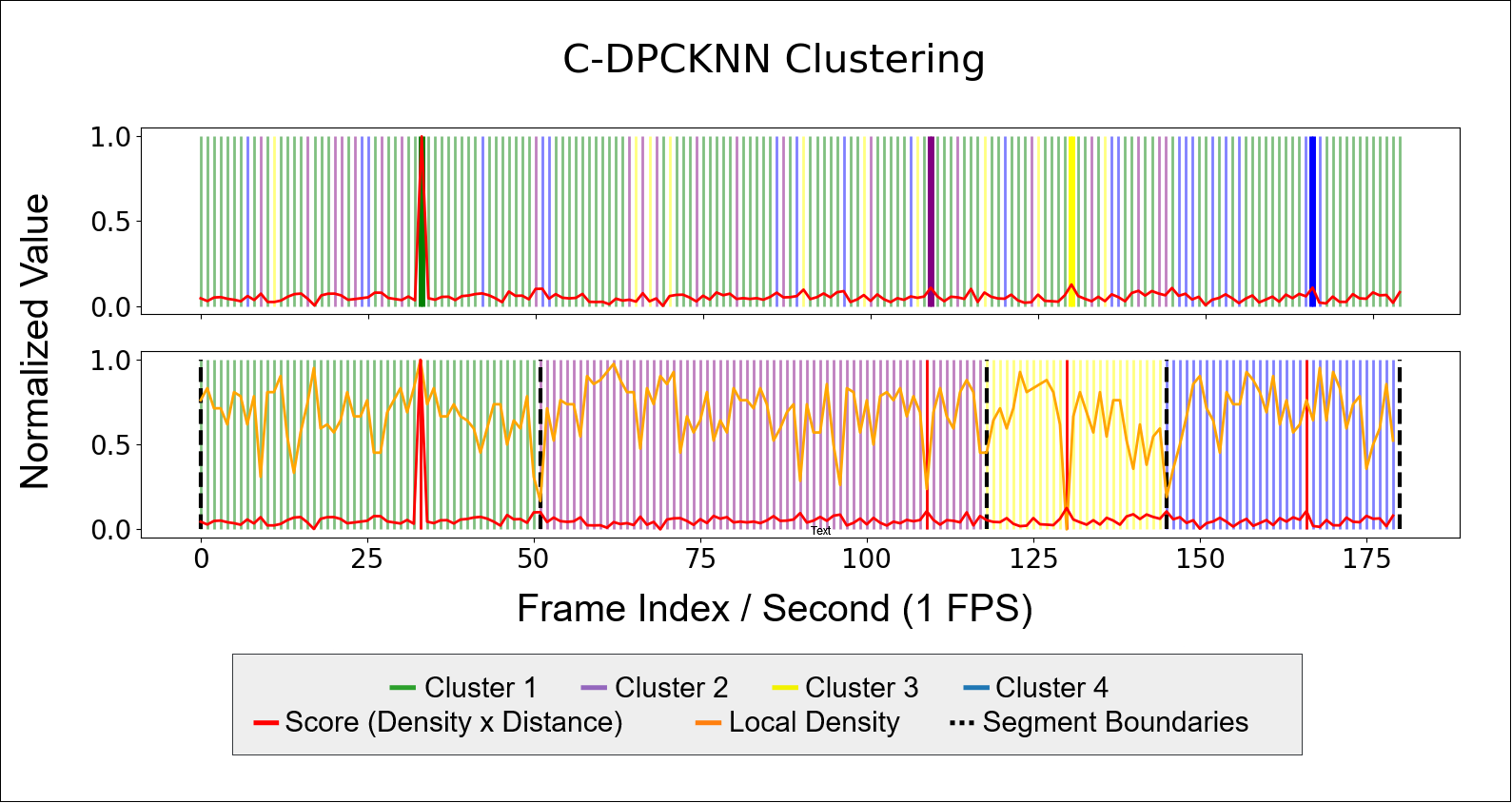}
     \caption{The Upper figure illustrates the intermediate results of DPCKNN clustering with blended boundaries among clusters.
     The bottom figure illustrates clearer event boundaries with the proposed C-DPCKNN.}
     \label{fig: TA-segment}
 \end{figure}

We visualize the temporal segmentation performance on EgoSchema. As seen in Figure \ref{fig: TA-segment}, the upper figure illustrates the intermediate clustering results with the original DPCKNN. According to the results, frames clustered to the same event are scattered across the video, and the event boundaries are blended, which is counter-intuitive to how untrimmed videos present their content. The bottom figure illustrates the results of how our proposed C-DPCKNN utilizes density peaks as sharp boundaries. This qualitative visualization shows that events are parsed correctly around clustering centers and the respective borders align to the regions with high fluctuations among frame features.

\footnotetext[2]{Obtained on the hidden test split of EgoSchema (5,000 tasks) instead of the public test split (500 tasks) as all the other results.}
\footnotetext[3]{Not obtained on the validation split of NExT-QA as the other results, but on the test split.}

\begin{figure}[t]
    \centering
    \includegraphics[width=1.0\linewidth]{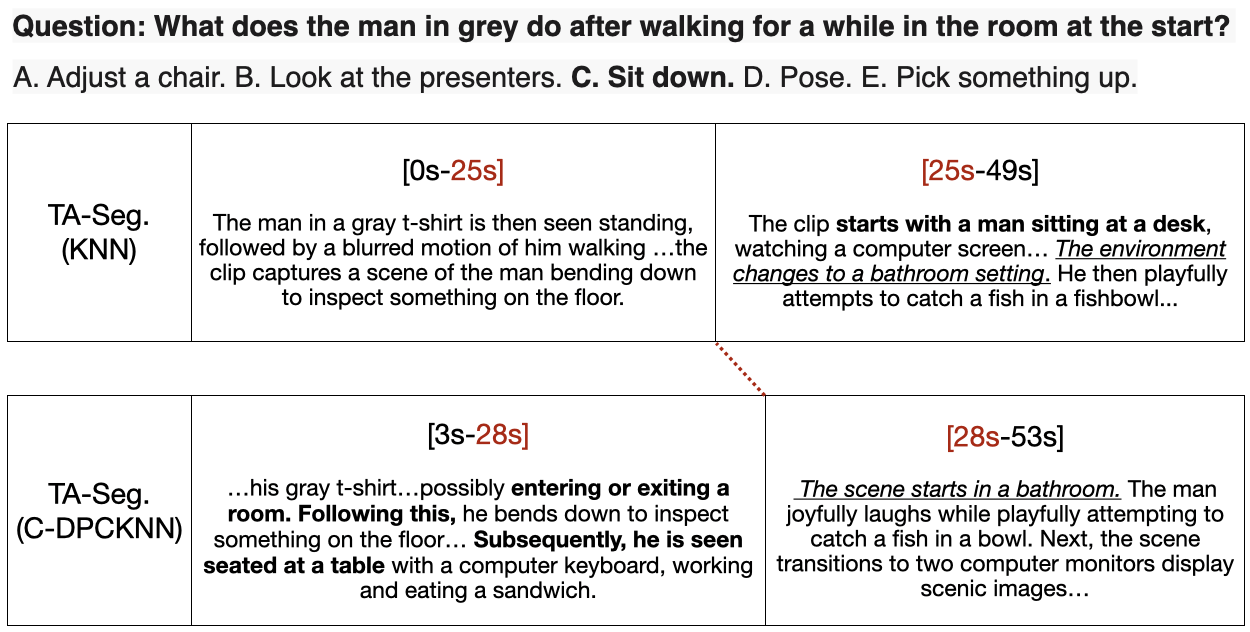}
    \caption{Performance of C-DPCKNN leads to clearer boundaries over KNN that contributes to exact semantic representations for videos segments.}
    \label{fig: quali-cdpcknn}
\end{figure}

\paragraph{Clear Segmentation for Correct Grounding} 
We further investigate how the two variants of VideoINSTA w/. TA-Seg(KNN) and TA-Seg(C-DPCKNN) affects the grounding descriptions.
We can find that the density-based clustering in C-DPCKNN successfully captures the scene transitions indicating the borders are set to where the content changes drastically, when the man starts to catch fish in a fishbowl in the bathroom as underlined in Figure \ref{fig: quali-cdpcknn}.
The consequent actions of the man in gray before he went to the bathroom are fully tracked in the same clip, leading to the correct answer “C) sit down”. However, the KNN method falsely sets the border causing important information loss leading to the false answer “E) pickup something”.

\paragraph{Spatially Informative Captions}
VLMs share a tendency to focus on describing the actions and events happning in the video clips or frames. However, the environment in videos and the interactions between human and objects provide more trivial but essential information for accurate reasoning in a fine-grained level, to which the spatially informative reasoning with object detections contributes. An example in IntentQA has the answer \textit{"Seat belt"} to the question \textit{"How did the people make sure that the babies will not fall off the swing easily when playing on them?"}. Basic video narrations will lead to captions like \textit{"Some people are standing around the babies and playing swings."}, leading to a false prediction of \textit{"Standing Around"}, while neglecting the crucial factor for safety, which actually is the object \textit{seat belt}.


\section{Conclusion}
This work focuses on understanding long-form videos with LLMs -- 
particularly emphasizing information quality, spatial-temporal reasoning, and explicit complex reasoning across unbalanced distributed information.
The proposed training-free framework \textbf{VideoINSTA} 
for long-form video understanding showcases exceeding performance over 
state-of-the-art end-to-end and zero-shot LLM-based methods. It further reveals the potential on open question answering 
and the extensibility of various visual-language tool-augmented spatial-temporal reasoning approaches.


\section*{Limitation}

The limitation of \textbf{VideoINSTA} lies in its nature as a compound system, centered around a large language model (LLM) and incorporating various visual-language tools to process spatial-temporal information. If the number of tools or the rounds of reasoning increase to some level, there is a heightened risk of inconsistency and randomness of generated intermediate thoughts, potentially introducing additional noise into the reasoning process.


\section*{Ethics Statement}



VideoINSTA is tailored as a compound system utilizing various visual-language tools for spatial-temporal information extraction. This framework might help with developing visual understanding systems for assisting daily life since it has exceeding results on the first-view dataset EgoSchema. 
The risk of VideoINSTA might be inherited from open-source LLMs, such as bias and hallucinations. Besides, We only use AI assistants (e.g., ChatGPT) to conduct experiments in this research.

\section*{Liscences}
The datasets used in this research work are open-sourced and can be seen in references. We use the datasets from the original version within the intended use term. The licenses of the models used in this paper are listed.

\begin{itemize}
    \item \href{https://github.com/CeeZh/LLoVi/blob/main/LICENSE}{LLoVi}
    \item \href{https://ego4d-data.org/pdfs/Ego4D-Licenses-Draft.pdf}{EgoSchema}
    \item \href{https://github.com/doc-doc/NExT-QA?tab=MIT-1-ov-file}{NExT-QA}
    \item \href{https://github.com/JoseponLee/IntentQA?tab=readme-ov-file}{Intent-QA}
    \item \href{https://github.com/showlab/UniVTG/blob/main/LICENSE}{UniVTG}
    \item \href{https://github.com/PKU-YuanGroup/Chat-UniVi/blob/main/LICENSE}{Chat-UniVi}
    \item \href{https://github.com/THUDM/CogAgent/blob/main/LICENSE}{CogAgent}
    \item \href{https://github.com/meta-llama/llama3/blob/main/LICENSE}{LLama3}
\end{itemize}


\section*{Acknowledgements}

This work was funded by the Munich Center for Machine Learning and supported by the Federal Ministry of Education and Research and the State of Bavaria. 

\bibliography{acl_latex}

\appendix


\definecolor{gray_lvl_1}{RGB}{153, 153, 153}

\clearpage

\section{Case Studies}
\paragraph{Success Case}

\begin{figure}[h]
    \centering
    \includegraphics[width=1.0\linewidth]{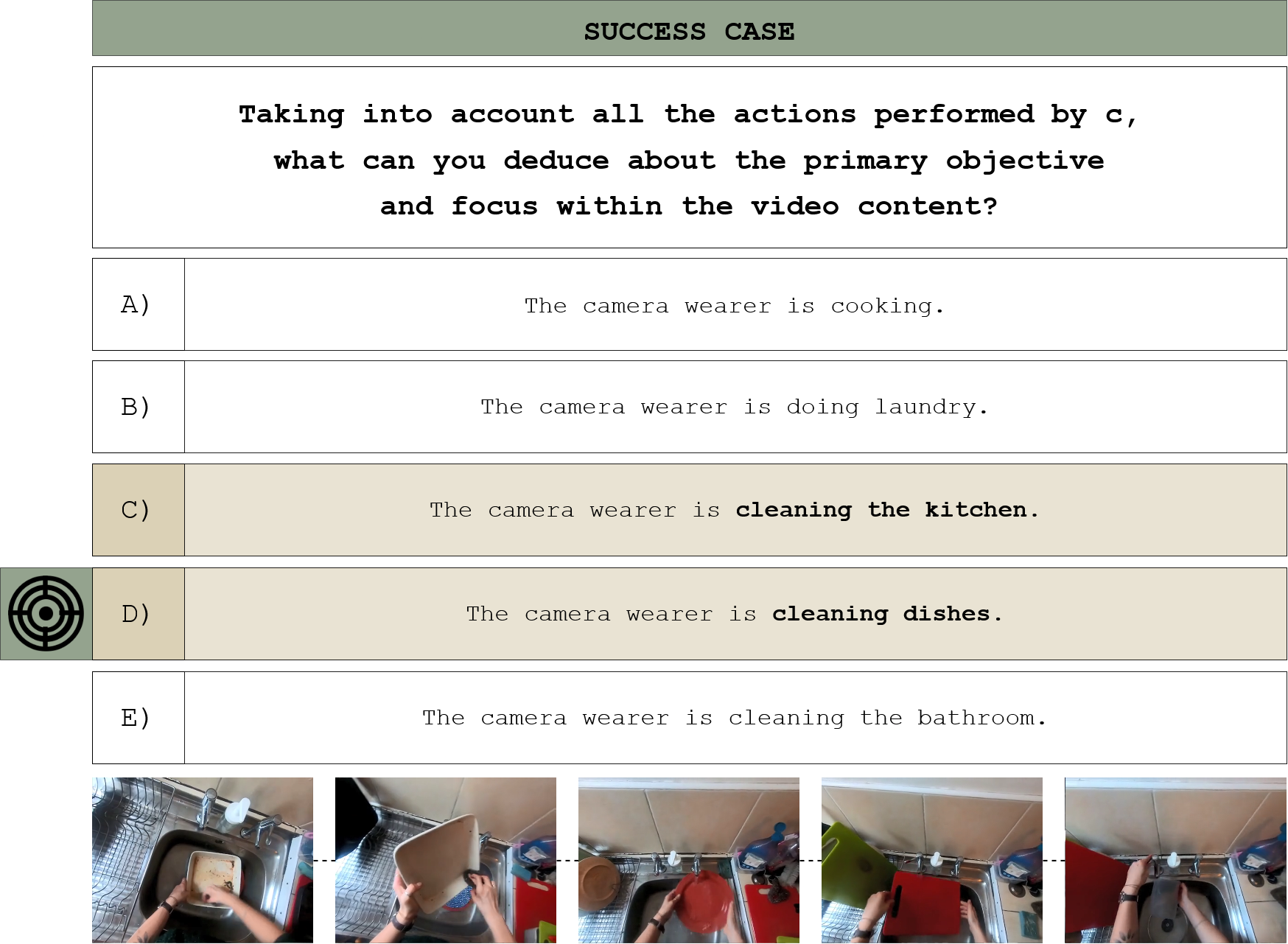}
    \caption{Sucess case. The ambiguity between the answer options C) and D) is highlighted in bold. The ground
    truth answer option is marked with a bullseye symbol and the prediction of the
    VideoINSTA framework is indicated with a crosshair symbol. In this case, they are
    overlapped.}
    \label{fig: sucess}
\end{figure}
As shown in Figure \ref{fig: sucess}, the VideoINSTA framework effectively addresses the ambiguity between the actions "cleaning dishes" and "cleaning the kitchen." While "cleaning the kitchen" appears broader and potentially applicable, "cleaning dishes" is more specific to the actual video content. A human viewer, after watching the video and reviewing the answer options, would likely determine that the individual is focused solely on cleaning dishes, rather than wiping kitchen surfaces or completing other tasks. Thus, "cleaning dishes" is the more accurate selection.

\begin{figure}[h]
    \centering
    \includegraphics[width=1.0\linewidth]{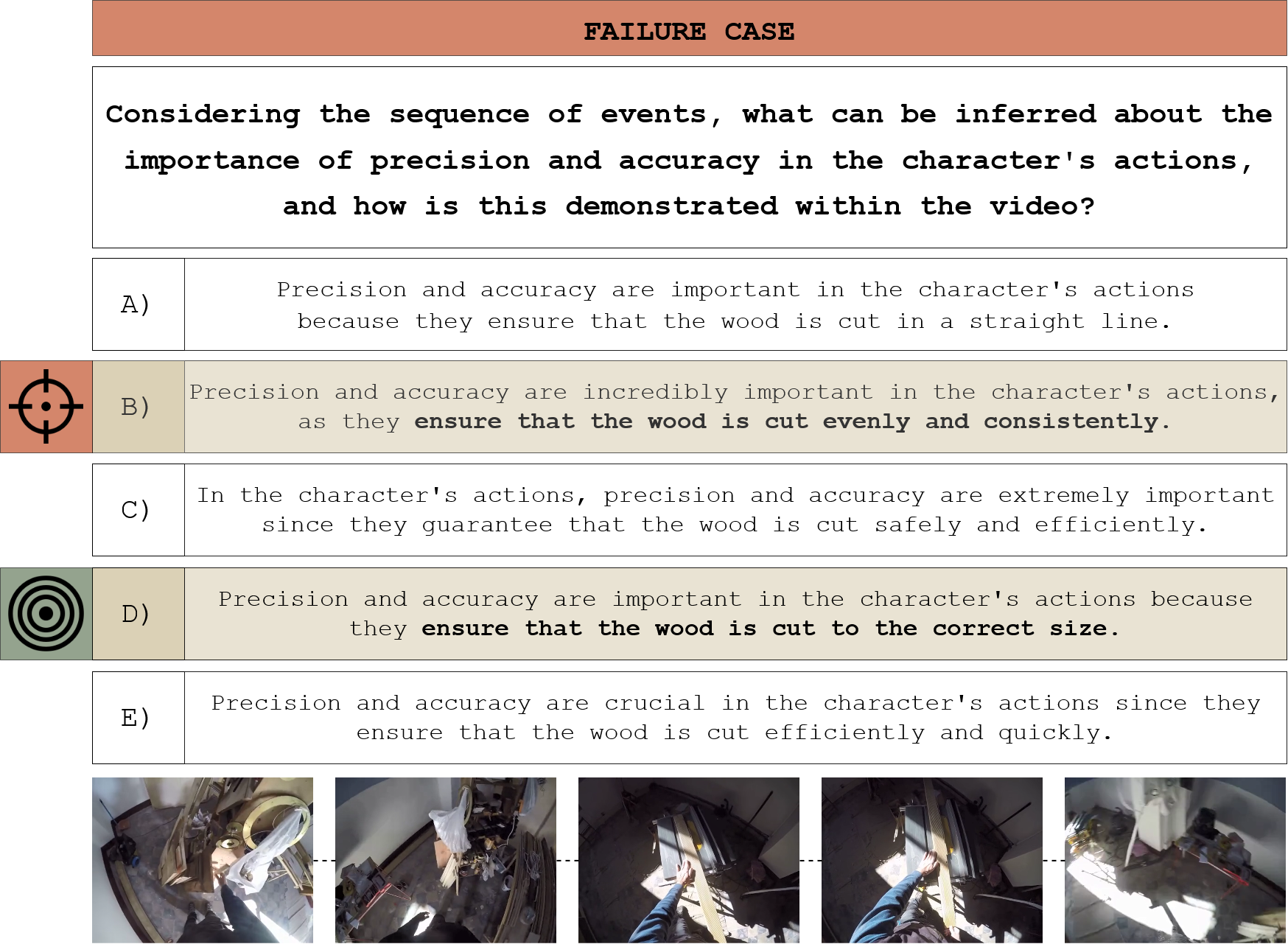}
    \caption{Failure case. The ambiguity
    between the answer options B) and D) is highlighted in bold. The ground truth
    answer option is marked with a bullseye symbol and the prediction of the VideoINSTA
    framework is indicated with a crosshair symbol. In this case, our algorithm
    fails to predict the ground truth option D and aims for B) instead.}
    \label{fig: failure}
\end{figure}

\paragraph{Failure Case}
Figure \ref{fig: failure} shows the failure case. The task is to determine whether the importance of precision stems from the need to cut the wood "evenly and consistently" (option B) or to the "correct size" (option D). A brief review of the video might suggest that both options are plausible. 
However, watching the full video reveals that only a single piece of wood is involved throughout, making "cutting to the correct size" the more accurate answer. The option of "cutting evenly and consistently" would imply the presence of multiple pieces, which is not the case, even when the wood temporarily leaves the camera's view. Unlike a human, who intuitively recognizes that the reappearing wood is still the same and that no other pieces exist, VideoINSTA struggles to track it consistently due to its lack of an environmental conciousness and the inability to track object identity. This shortcoming prevents VideoINSTA from recognizing that "cutting evenly and consistently" is irrelevant in this scenario, leading to the selection of an incorrect answer instead of the ground-truth response.

\section{More Related Works}

\paragraph{Video Language Models}
With the in-depth investigation into Multi-modal Large Language Models(MLLMs)~\cite{gu2023systematic, wu2023multimodal, cui2024survey}, there has been growing attention to bridging video modality to generative large language models such as Video-llama~\cite{zhang2023video}, Video-LLaVA~\cite{lin2023video}, LanguageBind~\cite{zhu2023languagebind}, VideoChat~\cite{2023videochat}, ChatUniV~\cite{Jin_2024_CVPR} InternVideo~\cite{wang2022internvideo}, etc., which are dependent on meticulously designed model structures, or adaptation layers. They suffer from large-
scale pertaining, or requiring proper datasets for instruction tuning such as InternVid~\cite{wang2023internvid}. Therefore, a line of work utilizing LLM as a compound system center or agent-based reasoning for video understanding has been introduced, which we discussed extensively in our baselines in Sec. \ref{sec: rel_works} and experiments \ref{tab: main_results}. Another line of work, focusing on low-resource and even zero-shot understanding of videos emerges, such as LLaVA-Next~\cite{liu2024llavanext}, E3M~\cite{bao2024e3m}, LongVLM~\cite{weng2024longvlm}, C2C~\cite{li2024c2c},~\cite{choi2024neuro}, etc, where they enlight the task more lightly. 

\section{Supplementary Statistics} \label{sss: ap_dataset stat}

\paragraph{Dataset Statistics} We report the split that we use for our experiments in Table \ref{tab: dataset}, the number of tasks in those splits -- i.e. the number of question-answer-pairs -- as well as the number of videos within those splits. Furthermore, we report the average, minimum and maximum video length in seconds of the videos in the corresponding split -- these numbers can vary from the ones for the whole datasets.
\begin{table}[!ht]
\begin{center}
\resizebox{0.48\textwidth}{!}{
    \centering
    \resizebox{0.6\textwidth}{!}{
    \begin{tabular}{c|c|c|c|c|c|c}
    \hline
        \cellcolor{gray_lvl_1} Datasets & \cellcolor{gray_lvl_1} Split & \cellcolor{gray_lvl_1} \#Tasks & \cellcolor{gray_lvl_1} \#Videos & \cellcolor{gray_lvl_1} Avg. Length & \cellcolor{gray_lvl_1} Min. Length & \cellcolor{gray_lvl_1} Max. Length \\ \hline
        EgoSchema & Public Test & 500 & 500 & 180.0 & 180.0 & 180.0 \\ \hline
        NExT-QA & Validation & 4,996 & 570 & 42.2 & 10.0 & 180.0 \\ \hline
        IntentQA & Test & 2,134 & 576 & 44.9 & 6.0 & 180.0 \\ \hline
        ActivityNet-QA & Test & 8,000 & 800 & 112.1 & 3.0 &  285.7 \\ \hline
    \end{tabular}
}}\end{center}
\caption{Dataset statistics.}
\label{tab: dataset}
\end{table}

\paragraph{Pre-trained model versions and statistics}  As shown in Table \ref{tab: models}, we abbreviate Large Language Model with LLM, Vision Language Model with VLM, Visual Temporal Grounding Model with VTGM, and Vision Encoder with VE. Please refer to the implementation details for the exact hyper-parameters that we use, since they vary for some different experiments and use cases.
\begin{table}[!ht]
\begin{center} 
\resizebox{0.47\textwidth}{!}{
    \centering
    \resizebox{0.6\textwidth}{!}{
    \begin{tabular}{c|c|c|c|c}
    \hline
        \cellcolor{gray_lvl_1} Models & \cellcolor{gray_lvl_1} Version & \cellcolor{gray_lvl_1} Type & \cellcolor{gray_lvl_1} \#Params & \cellcolor{gray_lvl_1} Context \\ \hline
        ChatGPT 3.5 & gpt-3.5-turbo-1106 & LLM & N/A & 16k \\ \hline
        ChatGPT 4 & gpt-4-1106-preview & LLM & N/A & 128k \\ \hline
        Llama3 & meta-llama/Meta-Llama-3-8B-Instruct & LLM & 8B & 8k \\ \hline
        UniVTG & CLIP-B/32 Pretraining (Finetuned) & VTGM & N/A & N/A \\ \hline
        LaViLa & Fair Checkpoint ~\citep{zhang2023simple} & VLM & N/A & 0 \\ \hline
        CogAgent & THUDM/cogagent-vqa-hf, lmsys/vicuna-7b-v1.5 & VLM & 18B & N/A \\ \hline
    \end{tabular}
}}\end{center}
\caption{Pre-trained model versions and statistics.}
\label{tab: models}
\end{table}

\begin{table}[!ht]
    \centering
    \resizebox{0.45\textwidth}{!}{
    \begin{tabular}{l|c|c}
    \hline
    \cellcolor{gray_lvl_1} Method & \cellcolor{gray_lvl_1} EgoSchema & \cellcolor{gray_lvl_1} NExT-QA \\ \hline
    w. TA-Seg. (Uniform) & 0.600 (±0.004) & 0.644 (±0.006) \\ \hline
    w. TA-Seg. (KNN) & 0.609 (±0.003) & 0.640 (±0.004) \\ \hline
    w. TA-Seg (DPCKNN) & 0.601(±0.001) & 0.647 (±0.001 \\ \hline
    w. TA-Seg. (C-DPCKNN) & \textbf{0.628 (±0.001)} & \textbf{0.679 (±0.009)} \\ \hline
    \end{tabular}}
    \caption{Ablation on different temporal segmentation of VideoINSTA methods on EgoSchema and NExT-QA datasets.}
    \label{tab:performance_metrics}
\end{table}

\section{Implementation Details}

\subsection{Experiment Setup}

We split a dataset into equal-sized chunks and run a sub-experiment on each of them for parallelization purposes. We collect and aggregate the results of all sub-experiments afterward to obtain the final experiment result. We use the types of GPU servers: NVIDIA RTX A6000 GPU, NVIDIA A100-PCIE-40GB, Quadro RTX 8000, NVIDIA RTX 3090. 

\subsubsection{Details of Llama3}

When we refer to Llama3, we use the instruction-tuned version \textit{meta-llama/Meta-Llama-3-8B-Instruct} ~\cite{llama3modelcard}, which is available on HuggingFace ~\cite{huggingface}. We use greedy sampling -- comparable with a temperature of $0.0$ -- throughout all our experiments. 

\subsubsection{Details of ChatGPT}

When we refer to ChatGPT 3.5, we use the instruction-tuned version \textit{gpt-3.5-turbo-1106}, and when we refer to ChatGPT 4, we use the instruction-tuned version \textit{gpt-4-1106-preview} ~\citealp{openai2024chatgpt, openaichatgptmodels}. Following ~\citep{zhang2023simple}, we use a temperature of $1.0$ for the summarization.

\subsubsection{Details of LaViLa}

For our experiments on EgoSchema, we use LaViLa ~\citep{zhao2023lavila} as the action captioner. Following ~\citep{zhang2023simple}, we use their retrained model checkpoint to avoid data leakage and ensure a fair comparison. We uniformly sample $4$ frames from each consecutive $1s$-interval of the video to obtain a caption. 

\subsubsection{Details of CogAgent}

Following ~\cite{wang2024videoagent}, we leverage the VLM CogAgent ~\cite{hong2024cogagent} as the action captioner for our experiments on NExT-QA, IntentQA and ActivityNetQA. Moreover, we use it as the label-free object detector for our experiments on all datasets. Specifically, we use the model \textit{THUDM/cogagent-vqa-hf} together with the tokenizer \textit{lmsys/vicuna-7b-v1.5}, which are available on HuggingFace ~\citep{huggingface}. 

\subsubsection{Details of UniVTG}

We leverage UniVTG ~\citep{lin2023univtg} to get the temporal grounding of a video and finally retrieve the most important interval regarding the question of a task. We use \textit{ViT-B/32} as the CLIP vision encoder model version ~\cite{radford2021learning} together with their best-fine-tuned model checkpoint ~\citep{univtg}. 

\subsubsection{Details of C-DPCKNN}

We use the CLIP vision encoder \textit{openai/clip-vit-large-patch14} ~\citep{radford2021learning}, which is available on HuggingFace ~\citep{huggingface}. 




\subsection{Prompts} \label{app: prompt}

\definecolor{coffee}{RGB}{247, 218, 171}
\definecolor{matcha}{RGB}{96, 128, 119}

\begin{table*}[t]
\centering
\begin{tabular}{|c|}
\hline
\begin{minipage}{0.9\textwidth}
\small
\cellcolor{coffee}
\texttt{\\
You are given some language descriptions of a first-person view video. The video is \textcolor{matcha}{\{length\}} seconds long. Each sentence describes a 1.0s clip. The descriptions are sequential and non-overlapping which cover the whole video exactly. Here are the descriptions: \textcolor{matcha}{\{interval\_text\}}.\textbackslash n Please give me a \textcolor{matcha}{\{words\}} words summary. When doing summarization, remember that your summary will be used to answer this multiple choice question: \textcolor{matcha}{\{question\}}
\\
}
\end{minipage} \\
\hline
\end{tabular}
\caption{Action Captions Summarization Prompt Template for ChatGPT. Note that only linebreaks explicitly indicated with "\textbackslash n" are true linebreaks at runtime -- the linebreaks of this document are just for more readability. Parameters being filled at runtime are indicated with \textcolor{matcha}{\{coloured single curly brackets\}}. }
\label{tab: action_cations_summarization_prompt_template_chatgpt}
\end{table*}

\begin{table*}[t]
\centering
\begin{tabular}{|c|}
\hline
\begin{minipage}{0.9\textwidth}
\small
\cellcolor{coffee}
\texttt{\\
You are given some language descriptions of a first person view video. The video is \textcolor{matcha}{\{length\}} seconds long. Each sentence describes a 1.0s clip. The descriptions are sequential and non-overlapping which cover the whole video exactly. Here are the descriptions: \textcolor{matcha}{\{interval\_text\}}.\textbackslash n \textbf{Please give me a summary of these action captions. Please write an easy-to-read continuous text. You can use paragraphs, but do not use special formatting such as bulleted or numbered lists. Please use \textcolor{matcha}{\{words\}} words for your summary.} When doing summarization, remember that your summary will be used to answer this multiple choice question: \textcolor{matcha}{\{question\}}
\\
}
\end{minipage} \\
\hline
\end{tabular}
\caption{Action Captions Summarization Prompt Template for Llama3. The difference to the prompt template for ChatGPT is highlighted in \textbf{bold}. Note that only linebreaks explicitly indicated with "\textbackslash n" are true linebreaks at runtime -- the linebreaks of this document are just for more readability. Parameters being filled at runtime are indicated with \textcolor{matcha}{\{coloured single curly brackets\}}.}
\label{tab: action_captions_summarization_prompt_template_llama3}
\end{table*}

\begin{table*}[t]
\centering
\begin{tabular}{|c|}
\hline
\begin{minipage}{0.9\textwidth}
\small
\cellcolor{coffee}
\texttt{\\
You are given a list of the most eye-catching objects that were detected in each frame of a video clip using a visual large language model. The list appears in the temporal order of the frames. The video is \textcolor{matcha}{\{length\}} seconds long. Each sentence describes the objects of a 1.0s clip. The object detections are sequential and non-overlapping which cover the whole video exactly. Here are the object detections:\textbackslash n\textbackslash n\textcolor{matcha}{\{interval\_text\}}.\textbackslash n\textbackslash nPlease give me a \textcolor{matcha}{\{words\}} words summary of these object detections. When doing summarization, remember that your summary will be used to answer this multiple choice question: \textcolor{matcha}{\{question\}}
\\
}
\end{minipage} \\
\hline
\end{tabular}
\caption{Object Detections Summarization Prompt Template for ChatGPT. Note that only linebreaks explicitly indicated with "\textbackslash n" are true linebreaks at runtime -- the linebreaks of this document are just for more readability. Parameters being filled at runtime are indicated with \textcolor{matcha}{\{coloured single curly brackets\}}.}
\label{tab: object_detections_summarization_prompt_template_chatgpt}
\end{table*}

\begin{table*}[t]
\centering
\begin{tabular}{|c|}
\hline
\begin{minipage}{0.9\textwidth}
\small
\cellcolor{coffee}
\texttt{\\
You are given a list of the most eye-catching objects that were detected in each frame of a video clip using a visual large language model. The list appears in the temporal order of the frames. The video is \textcolor{matcha}{\{length\}} seconds long. Each sentence describes the objects of a 1.0s clip. The object detections are sequential and non-overlapping which cover the whole video exactly. Here are the object detections:\textbackslash n\textbackslash n\textcolor{matcha}{\{interval\_text\}}.\textbackslash n\textbackslash n\textbf{Please give me a summary of these object detections. Please write an easy-to-read continuous text. You can use paragraphs, but do not use special formatting such as bulleted or numbered lists. Please use \textcolor{matcha}{\{words\}} words for your summary.} When doing summarization, remember that your summary will be used to answer this multiple choice question: \textcolor{matcha}{\{question\}}
\\
}
\end{minipage} \\
\hline
\end{tabular}
\caption{Object Detections Summarization Prompt Template for Llama3. The difference to the prompt template for ChatGPT is highlighted in \textbf{bold}. Note that only linebreaks explicitly indicated with "\textbackslash n" are true linebreaks at runtime -- the linebreaks of this document are just for more readability. Parameters being filled at runtime are indicated with \textcolor{matcha}{\{coloured single curly brackets\}}.}
\label{tab: object_detections_summarization_prompt_template_llama3}
\end{table*}

\begin{table*}[t]
\centering
\begin{tabular}{|c|}
\hline
\begin{minipage}{0.9\textwidth}
\small
\cellcolor{coffee}
\texttt{\\
\# Video Question Answering\\
\textbackslash n\textbackslash nHi there! Now that you have studied the topic of video question answering for years, you find yourself in the final exam of your studies. Please take your time to solve this task. You can do it! You know everything that is required to master it. Good luck!\\\\
\textbackslash n\textbackslash n\#\# What is Video Question Answering?\\
\textbackslash n\textbackslash nVideo Question Answering is a task that requires reasoning about the content of a video to answer a question about it. In this exam, you will be given purely textual information about a single clip of the video that has been extracted beforehand. Your task is to read the information about the clip carefully and evaluate whether the given clip is needed to answer the question about the video or not.\\\\
\textbackslash n\textbackslash n\#\# Here is your task\\
\textbackslash n\textbackslash nPlease think step by step to evaluate the answerability of the given question and options based on the given clip. The question is a single choice question with five answer options, such that there is exactly one best answer option. Is the information in the given clip sufficient to answer the given question with one of the given options? Please make sure to include all relevant information in your evaluation.\\\\
\textbackslash n\textbackslash nPlease use the following criteria for evaluation:\\
\textbackslash n\ \ \ \ 1. Irrelevant information \{\{'answerability': 1\}\}: If information of this clip is not even relevant to the question.\\
\textbackslash n\ \ \ \ 2. Insufficient information \{\{'answerability': 2\}\}: If information of this clip is potentially useful to answer the question, but more clips are needed to confidently answer the question.\\
\textbackslash n\ \ \ \ 3. Sufficient information \{\{'answerability': 3\}\}: If the information of this clip is sufficient to answer the question and no other clip is needed.\\\\
\textbackslash n\textbackslash nPlease write your answerability X in JSON format \{\{'answerability': X\}\}, where X is in \{\{1, 2, 3\}\}.\\\\
\textbackslash n\textbackslash n\#\# Here is the information about the video clip\\
\textbackslash n\textbackslash n\#\#\# Information about one of four clips of the video\\
\textbackslash n\textcolor{matcha}{\{lexical\_node\_state\_representation\}}\\\\
\textbackslash n\textbackslash n\#\#\# Question\\
\textbackslash n\textbackslash n\textcolor{matcha}{\{question\}}\\\\
\textbackslash n\textbackslash n\#\#\# Five answer options\\
\textbackslash n\textbackslash n\ \ \ \ A) \textcolor{matcha}{\{option\_0\}}\\
\textbackslash n\ \ \ \ B) \textcolor{matcha}{\{option\_1\}}\\
\textbackslash n\ \ \ \ C) \textcolor{matcha}{\{option\_2\}}\\
\textbackslash n\ \ \ \ D) \textcolor{matcha}{\{option\_3\}}\\
\textbackslash n\ \ \ \ E) \textcolor{matcha}{\{option\_4\}}\\\\
\textbackslash n\textbackslash n\#\# Now it is your turn\\
\textbackslash n\textbackslash nPlease think step by step to provide your evaluation and provide the answerability X in JSON format \{\{'answerability': X\}\}, where X is in \{\{1, 2, 3\}\}:\\
\textbackslash n\textbackslash n
\\
}
\end{minipage} \\
\hline
\end{tabular}
\caption{Answerability Rating Prompt Template for ChatGPT. Note that only linebreaks explicitly indicated with "\textbackslash n" are true linebreaks at runtime -- the linebreaks of this document are just for more readability. Parameters being filled at runtime are indicated with \textcolor{matcha}{\{coloured single curly brackets\}}. JSON-formatting is indicated by \{\{double curly brackets\}\}, as one level of brackets will be removed when the prompt template gets filled.}
\label{tab: answerability_rating_prompt_template_chatgpt}
\end{table*}

\begin{table*}[t]
\centering
\begin{tabular}{|c|}
\hline
\begin{minipage}{0.9\textwidth}
\small
\cellcolor{coffee}
\texttt{\\
\# Video Question Answering\\
\textbackslash n\textbackslash nHi there! Now that you have studied the topic of video question answering for years, you find yourself in the final exam of your studies. Please take your time to solve this task. You can do it! You know everything that is required to master it. Good luck!\\\\
\textbackslash n\textbackslash n\#\# What is Video Question Answering?\\
\textbackslash n\textbackslash nVideo Question Answering is a task that requires reasoning about the content of a video to answer a question about it. In this exam, you will be given purely textual information about a single clip of the video that has been extracted beforehand. Your task is to read the information about the clip carefully and evaluate whether the given clip is needed to answer the question about the video or not.\\\\
\textbackslash n\textbackslash n\#\# Here is your task\\
\textbackslash n\textbackslash nPlease think step by step to evaluate the answerability of the given question and options based on the given clip. The question is a single choice question with five answer options, such that there is exactly one best answer option. Is the information in the given clip sufficient to answer the given question with one of the given options? Please make sure to include all relevant information in your evaluation. \textbf{Moreover, make sure that you always provide an answerability, even if it seems ambiguous or unsolvable.}\\\\
\textbackslash n\textbackslash nPlease use the following criteria for evaluation:\\
\textbackslash n\ \ \ \ 1. Irrelevant information \{\{'answerability': 1\}\}: If information of this clip is not even relevant to the question.\\
\textbackslash n\ \ \ \ 2. Insufficient information \{\{'answerability': 2\}\}: If information of this clip is potentially useful to answer the question, but more clips are needed to confidently answer the question.\\
\textbackslash n\ \ \ \ 3. Sufficient information \{\{'answerability': 3\}\}: If the information of this clip is sufficient to answer the question and no other clip is needed.\\\\
\textbackslash n\textbackslash nPlease write your answerability X in JSON format \{\{'answerability': X\}\}, where X is in \{\{1, 2, 3\}\}.\\\\
\textbackslash n\textbackslash n\#\# Here is the information about the video clip\\
\textbackslash n\textbackslash n\#\#\# Information about one of four clips of the video\\
\textbackslash n\textcolor{matcha}{\{lexical\_node\_state\_representation\}}\\\\
\textbackslash n\textbackslash n\#\#\# Question\\
\textbackslash n\textbackslash n\textcolor{matcha}{\{question\}}\\\\
\textbackslash n\textbackslash n\#\#\# Five answer options\\
\textbackslash n\textbackslash n\ \ \ \ A) \textcolor{matcha}{\{option\_0\}}\\
\textbackslash n\ \ \ \ B) \textcolor{matcha}{\{option\_1\}}\\
\textbackslash n\ \ \ \ C) \textcolor{matcha}{\{option\_2\}}\\
\textbackslash n\ \ \ \ D) \textcolor{matcha}{\{option\_3\}}\\
\textbackslash n\ \ \ \ E) \textcolor{matcha}{\{option\_4\}}\\\\
\textbackslash n\textbackslash n\#\# Now it is your turn\\
\textbackslash n\textbackslash nPlease think step by step to provide your evaluation and provide the answerability X in JSON format \{\{'answerability': X\}\}, where X is in \{\{1, 2, 3\}\}:\\
\textbackslash n\textbackslash n
\\
}
\end{minipage} \\
\hline
\end{tabular}
\caption{Answerability Rating Prompt Template for Llama3. The difference to the prompt template for ChatGPT is highlighted in \textbf{bold}. Note that only linebreaks explicitly indicated with "\textbackslash n" are true linebreaks at runtime -- the linebreaks of this document are just for more readability. Parameters being filled at runtime are indicated with \textcolor{matcha}{\{coloured single curly brackets\}}. JSON-formatting is indicated by \{\{double curly brackets\}\}, as one level of brackets will be removed when the prompt template gets filled.}
\label{tab: answerability_rating_prompt_template_llama3}
\end{table*}

\begin{table*}[t]
\centering
\begin{tabular}{|c|}
\hline
\begin{minipage}{0.9\textwidth}
\small
\cellcolor{coffee}
\texttt{\\
\# Video Question Answering\\
\textbackslash n\textbackslash nHi there! Now that you have studied the topic of video question answering for years, you find yourself in the final exam of your studies. Please take your time to solve this task. You can do it! You know everything that is required to master it. Good luck!\\\\
\textbackslash n\textbackslash n\#\# What is Video Question Answering?\\
\textbackslash n\textbackslash nVideo Question Answering is a task that requires reasoning about the content of a video to answer a question about it. In this exam, you will be given purely textual information about one or more clips of a video that has been extracted beforehand. So your task is to read the information about the video carefully and answer the question about it.\\\\
\textbackslash n\textbackslash n\#\# Here is your task\\
\textbackslash n\textbackslash nBased on the given information about the most relevant clips of the video regarding the question, please select exactly one of the given options as your best answer to the given question. This is a single choice setting, such that there is exactly one best answer option. Think step by step to find the best candidate from the given answer options regarding the given question. Please write the letter of the best answer X in JSON format \{\{'best\_answer': 'X'\}\}, where X is in \{\{'A', 'B', 'C', 'D', 'E'\}\}.\\\\
\textbackslash n\textbackslash n\#\# Here is the information about the video\\
\textbackslash n\textbackslash n\#\#\# Information about the most relevant clips of the video regarding the question\\
\textbackslash n\textcolor{matcha}{\{whole\_video\_state\}}\\\\
\textbackslash n\textbackslash n\#\#\# Question\\
\textbackslash n\textbackslash n\textcolor{matcha}{\{question\}}\\\\
\textbackslash n\textbackslash n\#\#\# Five answer options (please select exactly one)\\
\textbackslash n\textbackslash n\ \ \ \ A) \textcolor{matcha}{\{option\_0\}}\\
\textbackslash n\ \ \ \ B) \textcolor{matcha}{\{option\_1\}}\\
\textbackslash n\ \ \ \ C) \textcolor{matcha}{\{option\_2\}}\\
\textbackslash n\ \ \ \ D) \textcolor{matcha}{\{option\_3\}}\\
\textbackslash n\ \ \ \ E) \textcolor{matcha}{\{option\_4\}}\\\\
\textbackslash n\textbackslash n\#\# Now it is your turn\\
\textbackslash n\textbackslash nPlease choose the best option now. Think step by step and provide the best answer (friendly reminder: in the requested JSON format \{\{'best\_answer': 'X'\}\}, where X is in \{\{'A', 'B', 'C', 'D', 'E'\}\}):\\
\textbackslash n\textbackslash n
\\
}
\end{minipage} \\
\hline
\end{tabular}
\caption{Question Answering Prompt Template for ChatGPT. Note that only linebreaks explicitly indicated with "\textbackslash n" are true linebreaks at runtime -- the linebreaks of this document are just for more readability. Parameters being filled at runtime are indicated with \textcolor{matcha}{\{coloured single curly brackets\}}. JSON-formatting is indicated by \{\{double curly brackets\}\}, as one level of brackets will be removed when the prompt template gets filled.}
\label{tab: question_answering_prompt_template_chatgpt}
\end{table*}

\begin{table*}[t]
\centering
\begin{tabular}{|c|}
\hline
\begin{minipage}{0.9\textwidth}
\small
\cellcolor{coffee}
\texttt{\\
\# Video Question Answering\\
\textbackslash n\textbackslash nHi there! Now that you have studied the topic of video question answering for years, you find yourself in the final exam of your studies. Please take your time to solve this task. You can do it! You know everything that is required to master it. Good luck!\\
\textbackslash n\textbackslash n\#\# What is Video Question Answering?\\\\
\textbackslash n\textbackslash nVideo Question Answering is a task that requires reasoning about the content of a video to answer a question about it. In this exam, you will be given purely textual information about one or more clips of a video that has been extracted beforehand. So your task is to read the information about the video carefully and answer the question about it.\\\\
\textbackslash n\textbackslash n\#\# Here is your task\\
\textbackslash n\textbackslash nBased on the given information about the most relevant clips of the video regarding the question, please select exactly one of the given options as your best answer to the given question. This is a single choice setting, such that there is exactly one best answer option. Think step by step to find the best candidate from the given answer options regarding the given question. Please write the letter of the best answer X in JSON format \{\{'best\_answer': 'X'\}\}, where X is in \{\{'A', 'B', 'C', 'D', 'E'\}\}. \textbf{Make sure that you always select the best answer option, even if it seems ambiguous or unsolvable.}\\\\
\textbackslash n\textbackslash n\#\# Here is the information about the video\\
\textbackslash n\textbackslash n\#\#\# Information about the most relevant clips of the video regarding the question\\
\textbackslash n\textcolor{matcha}{\{whole\_video\_state\}}\\\\
\textbackslash n\textbackslash n\#\#\# Question\\
\textbackslash n\textbackslash n\textcolor{matcha}{\{question\}}\\\\
\textbackslash n\textbackslash n\#\#\# Five answer options (please select exactly one)\\
\textbackslash n\textbackslash n\ \ \ \ A) \textcolor{matcha}{\{option\_0\}}\\
\textbackslash n\ \ \ \ B) \textcolor{matcha}{\{option\_1\}}\\
\textbackslash n\ \ \ \ C) \textcolor{matcha}{\{option\_2\}}\\
\textbackslash n\ \ \ \ D) \textcolor{matcha}{\{option\_3\}}\\
\textbackslash n\ \ \ \ E) \textcolor{matcha}{\{option\_4\}}\\\\
\textbackslash n\textbackslash n\#\# Now it is your turn\\
\textbackslash n\textbackslash nPlease choose the best option now. Think step by step and provide the best answer (friendly reminder: in the requested JSON format \{\{'best\_answer': 'X'\}\}, where X is in \{\{'A', 'B', 'C', 'D', 'E'\}\}):\\
\textbackslash n\textbackslash n
\\
}
\end{minipage} \\
\hline
\end{tabular}
\caption{Question Answering Prompt Template for Llama3. The difference to the prompt template for ChatGPT is highlighted in \textbf{bold}. Note that only linebreaks explicitly indicated with "\textbackslash n" are true linebreaks at runtime -- the linebreaks of this document are just for more readability. Parameters being filled at runtime are indicated with \textcolor{matcha}{\{coloured single curly brackets\}}. JSON-formatting is indicated by \{\{double curly brackets\}\}, as one level of brackets will be removed when the prompt template gets filled.}
\label{tab: question_answering_prompt_template_llama3}
\end{table*}

\begin{table*}[t]
\centering
\begin{tabular}{|c|}
\hline
\begin{minipage}{0.9\textwidth}
\small
\cellcolor{coffee}
\texttt{\\
\# Assessment of Decision-Making\\
\textbackslash n\textbackslash nHi there! You are given an exam task and a students answer to the task.\\
\textbackslash nYou are asked to assess the confidence level of the decision-making process in your students answer based on the information provided in the exam task. Imagine you are the teacher of the student and you want to know if you have provided enough information in the task to make a well-informed decision. At the same time, you want to know if the student has made a well-informed decision based on the information provided in the task.\\\\
\textbackslash n\textbackslash n\#\# Here is the exam\\
\textbackslash n\textbackslash n\textcolor{matcha}{\{reasoning\_history\}}\\\\
\textbackslash n\textbackslash n\#\# Criteria for Evaluation\\
\textbackslash n\textbackslash n\ \ \ \ 1. Insufficient Information \{\{'confidence': 1\}\}: If information is too lacking for a reasonable conclusion.\\
\textbackslash n\ \ \ \ 2. Partial Information \{\{'confidence': 2\}\}: If information partially supports an informed guess.\\
\textbackslash n\ \ \ \ 3. Sufficient Information \{\{'confidence': 3\}\}: If information fully supports a well-informed decision.\\\\
\textbackslash n\textbackslash n\#\# Assessment Focus\\
\textbackslash nPlease evaluate based on the relevance, completeness, and clarity of the provided information in the task in relation 
to the decision-making context of the students answer.\textbackslash nPlease provide the confidence in JSON format \{\{'confidence': X\}\} where X is in \{\{1, 2, 3\}\}.\textbackslash n\textbackslash n
\\
}
\end{minipage} \\
\hline
\end{tabular}
\caption{Self-Reflection Prompt Template for ChatGPT. Note that only linebreaks explicitly indicated with "\textbackslash n" are true linebreaks at runtime -- the linebreaks of this document are just for more readability. Parameters being filled at runtime are indicated with \textcolor{matcha}{\{coloured single curly brackets\}}. JSON-formatting is indicated by \{\{double curly brackets\}\}, as one level of brackets will be removed when the prompt template gets filled.}
\label{tab: self_reflection_prompt_template_chatgpt}
\end{table*}

\begin{table*}[t]
\centering
\begin{tabular}{|c|}
\hline
\begin{minipage}{0.9\textwidth}
\small
\cellcolor{coffee}
\texttt{\\
\# Assessment of Decision-Making\\
\textbackslash n\textbackslash nHi there! You are given an exam task and a students answer to the task.\\
\textbackslash nYou are asked to assess the confidence level of the decision-making process in your students answer based on the information provided in the exam task. Imagine you are the teacher of the student and you want to know if you have provided enough information in the task to make a well-informed decision. At the same time, you want to know if the student has made a well-informed decision based on the information provided in the task.\\\\
\textbackslash n\textbackslash n\#\# Here is the exam\\
\textbackslash n\textbackslash n\textcolor{matcha}{\{reasoning\_history\}}\\\\
\textbackslash n\textbackslash n\#\# Criteria for Evaluation\\
\textbackslash n\textbackslash n\ \ \ \ 1. Insufficient Information \{\{'confidence': 1\}\}: If information is too lacking for a reasonable conclusion.\\
\textbackslash n\ \ \ \ 2. Partial Information \{\{'confidence': 2\}\}: If information partially supports an informed guess.\\
\textbackslash n\ \ \ \ 3. Sufficient Information \{\{'confidence': 3\}\}: If information fully supports a well-informed decision.\\\\
\textbackslash n\textbackslash n\#\# Assessment Focus\\
\textbackslash nPlease evaluate based on the relevance, completeness, and clarity of the provided information in the task in relation 
to the decision-making context of the students answer.\textbackslash n\textbf{Please make sure that you always provide a confidence, even if it seems ambiguous or unsolvable.} Please provide the confidence in JSON format \{\{'confidence': X\}\} where X is in \{\{1, 2, 3\}\}.\textbackslash n\textbackslash n
\\
}
\end{minipage} \\
\hline
\end{tabular}
\caption{Self-Reflection Prompt Template for Llama3. The difference to the prompt template for ChatGPT is highlighted in \textbf{bold}. Note that only linebreaks explicitly indicated with "\textbackslash n" are true linebreaks at runtime -- the linebreaks of this document are just for more readability. Parameters being filled at runtime are indicated with \textcolor{matcha}{\{coloured single curly brackets\}}. JSON-formatting is indicated by \{\{double curly brackets\}\}, as one level of brackets will be removed when the prompt template gets filled.}
\label{tab: self_reflection_prompt_template_llama3}
\end{table*}

\begin{table*}[t]
\centering
\begin{tabular}{|c|}
\hline
\begin{minipage}{0.9\textwidth}
\small
\cellcolor{coffee}
\texttt{\\
You are given some language descriptions of a first person view video. The video is 63 seconds long. Each sentence describes a 1.0s clip. The descriptions are sequential and non-overlapping which cover the whole video exactly. Here are the descriptions: The camera wearer pours the water in the. The camera wearer picks a. The camera wearer washes the plate. The camera wearer washes the. The camera wearer washes the. The camera wearer scrapes the container. The camera wearer washes the plate. The camera wearer washes the. The camera wearer washes the. The camera wearer washes the tray with the sponge. The camera wearer washes the. The camera wearer washes the. The camera wearer washes the. The camera wearer washes the spoon. The camera wearer picks a. The camera wearer picks the bowl. The camera wearer washes the tray. The camera wearer washes the. The camera wearer washes the. The camera wearer washes the bowl. The camera wearer washes the. The camera wearer washes the. The camera wearer washes the. The camera wearer washes the. The camera wearer washes the tray. The camera wearer rinses the. The camera wearer pours water in the. The camera wearer rinses the. The camera wearer washes the tray. The camera wearer washes the. The camera wearer rinses the tray. The camera wearer closes the. The camera wearer lifts the basin. The camera wearer holds the tray with both. The camera wearer washes the. The camera wearer opens the. The camera wearer washes the tray with the sponge. The camera wearer washes the tray with the. The camera wearer closes the. The camera wearer holds the tray. The camera wearer opens the container. The camera wearer scrubs the. The camera wearer scrubs the sink. The camera wearer scrubs the sponge with a sponge scrub. The camera wearer scrubs the. The camera wearer scrubs the. The camera wearer scrubs the tray with a. The camera wearer wipes the board with a sponge. The camera wearer scrubs the board with a. The camera wearer squeezes the sponge. The camera wearer washes the chopping board. The camera wearer scrubs the chopping board with a. The camera wearer washes the chopping board. The camera wearer washes the chopping board with the. The camera wearer washes the. The camera wearer rinses chopping board. The camera wearer washes the chopping board. The camera wearer rinses the chopping. The camera wearer washes the chopping board. The camera wearer rinses the sponge. The camera wearer washes the chopping board with the. The camera wearer opens the sink. The camera wearer closes the dish. \\
 Please give me a 180 words summary. When doing summarization, remember that your summary will be used to answer this multiple choice question: Taking into account all the actions performed by the camera wearer, what can you deduce about the primary objective and focus within the video content?
\\
}
\end{minipage} \\
\hline
\end{tabular}
\caption{Action Caption Summarization Prompt Example for ChatGPT.}
\label{tab: action_cations_summarization_prompt_chatgpt}
\end{table*}

\begin{table*}[t]
\centering
\begin{tabular}{|c|}
\hline
\begin{minipage}{0.9\textwidth}
\small
\cellcolor{coffee}
\texttt{\\
You are given a list of the most eye-catching objects that were detected in each frame of a video clip using a visual large language model. The list appears in the temporal order of the frames. The video is 63 seconds long. Each sentence describes the objects of a 1.0s clip. The object detections are sequential and non-overlapping which cover the whole video exactly. Here are the object detections: 
\\ \\
Sink; Dish rack; Square dish. Sink; Dishwashing soap dispenser; Dish rack. Sink; Dish soap dispenser; Dish rack. Sink; Soap dispenser; Plastic bottle. Sink; Hand; Pan. Sink; Dish soap dispenser; Black pan. Sink; Dish soap dispenser; Plastic bottle. Sink; Dish soap dispenser; Plastic container. Sink; Hand; Dish soap. Sink; Dishwashing spray bottle; Dish rack. A sink; A dish rack; A person's hands. A sink; A faucet; A dish rack. Sink; Dishwashing soap dispenser; Dish rack. Sink; Dish rack; Soap dispenser. Sink; Plate with food remnants; Hand. Sink; Cutting board; Spray bottle. A sink; A hand washing dish soap dispenser; A red chopping board. A sink; A faucet; A spray bottle. A sink; A faucet; A bottle of dish soap. A sink; A black dish or container; A red cutting board. Sink; Dish soap dispenser; Plastic bottle. Sink; Hand; Plastic bottle. A sink; A faucet; A bottle of dish soap. Sink; Dish soap dispenser; Cutting board. Sink; Hands; Plastic bottle. Sink; Dishwashing soap dispenser; Plastic bottle. A black tray or dish; A white container or bowl; A bottle of liquid soap. Sink; Faucet; Dishwashing soap dispenser. Sink; Faucet; Dishwashing soap. A sink; A faucet; A dish rack. A black container; A white container; A faucet. A sink; A faucet; A black object (possibly a pan or a lid). A black plate; A silver dish rack; A silver sink with a faucet. A sink; A faucet; A dishwashing soap dispenser. A sink; A faucet; A dish rack. Sink; Plate; Cleaning spray bottle. Sink; Plate; Cleaning spray bottle. Sink; Plate; Dish soap. A sink; A white plate; A bottle of liquid. A white plate; A sink; A bottle. A green lid or cover; A red cutting board; A black container or pot. A white plate; A red cutting board; A bottle of cleaning solution. Plate; Sink; Dish rack. Sink; Plate; Dish rack. Sink; Dish rack; Plastic container. A white plate or dish; A metal dish rack; A sink. Sink; Dishwashing detergent bottle; Cutting board. A sink; A plate or tray; A bottle of dish soap. Sink; Plate; Cleaning bottle. A plate; A sink; A bottle of dish soap. A sink; A faucet; A bottle of dish soap. A sink; A dish rack; A bottle of dish soap. A sink; A dish rack; A bottle of dish soap. A sink; A dish rack; A bottle of dish soap. Sink; Plate; Cutting board. Sink; Plate; Soap dispenser. Sink; Plate; Dish soap dispenser. Sink; Plate; Dish soap. Sink; Plate; Soap dispenser. A sink; A dish rack; A bottle of dish soap. Sink; Plate; Dish soap. Sink; Dish soap dispenser; Red cutting board. A green container with a lid; A black frying pan or skillet; A metal dish rack. 
\\ \\
Please give me a 180 words summary of these object detections. When doing summarization, remember that your summary will be used to answer this multiple choice question: Taking into account all the actions performed by the camera wearer, what can you deduce about the primary objective and focus within the video content?
\\
}
\end{minipage} \\
\hline
\end{tabular}
\caption{Action Caption Summarization Prompt Example for ChatGPT.}
\label{tab: action_cations_summarization_prompt_chatgpt2} 
\end{table*}

\end{document}